\documentclass[10pt,twocolumn,letterpaper]{article}

\usepackage[final]{arxiv}              % To produce the CAMERA-READY version
\usepackage{multirow}
\usepackage{colortbl}
\usepackage{wrapfig}
\usepackage{graphicx}        % for \resizebox
\usepackage{booktabs}        % for \cmidrule, \toprule, \midrule, \bottomrule
\usepackage[table]{xcolor} 
\usepackage[most]{tcolorbox}
\newtcolorbox{promptbox}[2][]{
  colback=gray!5,
  colframe=gray!40,
  title={\textbf{#2}},
  fonttitle=\bfseries,
  left=1mm, right=1mm, top=1mm, bottom=1mm,
  enhanced,
  sharp corners,
  #1
}
\definecolor{red4}{HTML}{febf92}
\definecolor{blue4}{HTML}{77b1e2}
\definecolor{green4}{HTML}{F5FFFA}
\usepackage{xcolor} 
\definecolor{cvprblue}{rgb}{0.21,0.49,0.74}
\usepackage[pagebackref,breaklinks,colorlinks,allcolors=cvprblue]{hyperref}

\def\paperID{7696} % *** Enter the Paper ID here
\def\confName{CVPR}
\def\confYear{2026}

\title{Monet: Reasoning in Latent Visual Space Beyond Images and Language}

\author{
    Qixun Wang$^{1}$ \,\,
    Yang Shi$^{1,2}$ \,\,
    Yifei Wang$^{3}$\thanks{The work was done at MIT prior to joining Amazon.} \,\,
    \\
    Yuanxing Zhang$^{2}$ \,\,
    Pengfei Wan$^{2}$ \,\,
    Kun Gai$^{2}$ \,\,
    Xianghua Ying$^{1}$\thanks{Corresponding Author} \,\,
    Yisen Wang$^{1}$\footnotemark[2] \,\,
    \\
    $^{1}$Peking University \,\,
    $^{2}$Kling Team \,\,
    $^{3}$Amazon AGI SF Lab
}

\begin{document}
\maketitle

\begin{abstract}
``Thinking with images'' has emerged as an effective paradigm for advancing visual reasoning, extending beyond text-only chains of thought by injecting visual evidence into intermediate reasoning steps. However, existing methods fall short of human-like abstract visual thinking, as their flexibility is fundamentally limited by external tools. In this work, we introduce \textbf{Monet}, a training framework that enables multimodal large language models (MLLMs) to reason directly within the latent visual space by generating continuous embeddings that function as intermediate visual thoughts. We identify two core challenges in training MLLMs for latent visual reasoning—high computational cost in latent–vision alignment and insufficient supervision over latent embeddings, and address them with a three-stage distillation-based supervised fine-tuning (SFT) pipeline. We further reveal a limitation of applying GRPO to latent reasoning: it primarily enhances text-based reasoning rather than latent reasoning. To overcome this, we propose \textbf{VLPO} (\underline{\textbf{V}}isual-\underline{\textbf{l}}atent \underline{\textbf{P}}olicy \underline{\textbf{O}}ptimization), a reinforcement learning method that explicitly incorporates latent embeddings into policy gradient updates.
To support SFT, we construct Monet-SFT-125K, a high-quality text–image interleaved CoT dataset containing 125K real-world, chart, OCR, and geometry CoTs. Our model, Monet-7B, shows consistent gains across real-world perception and reasoning benchmarks and exhibits strong out-of-distribution generalization on challenging abstract visual reasoning tasks. We also empirically analyze the role of each training component and discuss our early unsuccessful attempts, providing insights for future developments in visual latent reasoning. Our model, data, and code are available at \url{https://github.com/NOVAglow646/Monet}.

\end{abstract}    
\section{Introduction}
\label{sec:intro}
Recent work \cite{hu2024visual, zheng2025deepeyes, fan2025grit, Realunify, sarch2025grounded} has demonstrated that the incorporation of auxiliary images in the intermediate steps of chain-of-thought (CoT) can improve the visual reasoning of multimodal large language models (MLLMs)~\cite{comanici2025gemini,bai2025qwen2,wang2025internvl3,mavors, cheng2025visual, jian2025look}. 
%Empirically, such auxiliary images improve visual reasoning through the following aspects: (1) re-emphasizing the visual features of crucial image regions by cropping or drawing bounding boxes, alleviating the gradual loss of attention to image tokens during reasoning \citep{cheng2025visual, jian2025look}. (2) serving as visualizations of the intermediate steps of the visual-related thoughts, such as auxiliary lines in a geometry problem and intermediate states in a maze problem. 
A recent trend is to acquire auxiliary images by training MLLMs to predict key region coordinates \citep{zheng2025deepeyes, fu2025refocus, su2025pixel, fan2025grit}, invoking visual tools such as grounding or depth estimation models \citep{hu2024visual, sarch2025grounded, su2025openthinkimg}, or generating executable codes \citep{zhao2025pyvision, zhang2025thyme} to modify the input image. While promising, these methods are constrained by a limited set of external tools, lacking the flexible, human-like visual reasoning that arises within an internal perceptual space. 

To emulate the flexible visual reasoning of humans, we investigate training MLLMs to reason directly in the continuous latent space. Specifically, we enable MLLMs to generate latent embeddings that serve as intermediate visual thoughts beyond textual descriptions and image embeddings, thus eliminating the need for explicit auxiliary images and overcoming the rigidity of external tools. Recent studies have begun exploring latent visual reasoning strategies \citep{yang2025machine, li2025latent, pham2025multimodal}. Most existing methods simply align the generated embeddings with those of auxiliary images while applying a next-token prediction (NTP) loss on text tokens during supervised fine-tuning (SFT), and directly apply GRPO \citep{Shao2024DeepSeekMathPT} for reinforcement learning (RL), which leads to two key limitations: \textbf{(1) poor scalability}, as alignment over thousands of image tokens incurs high computational and memory costs, and using mean pooling to compress the image tokens \citep{yang2025machine} will distort detailed visual features; \textbf{(2) insufficient optimization of latent embeddings}, since the NTP objective in SFT can be easily overfit and the GRPO loss can only be computed for text tokens, the optimization of latent embeddings are ignored; Consequently, their improvements remain limited and task-specific.

Motivated by these observations, we propose \textbf{Monet}\footnote{The name \textbf{Monet} is derived from the great Impressionist artist, as our method enables MLLMs to think abstractly in the latent visual space.}, a novel training framework for multimodal latent reasoning  that trains a text-output MLLM (Qwen2.5-VL-7B \citep{bai2025qwen2}) to perform latent reasoning through SFT and RL. The SFT stage aims at equipping the model with the fundamental ability to generate and reason with latent embeddings. To tackle limitation \textbf{(1)}, instead of directly aligning latent embeddings with those of auxiliary images, we introduce dual supervision signals. First, since latent embeddings are intended to facilitate reasoning, we align the hidden representations of text tokens corresponding to crucial visual features observed from the auxiliary images, when conditioned on either auxiliary images or generated latent embeddings. Second, to preserve visual information, we use a controlled attention mask so that latent embeddings can directly attend to auxiliary image embeddings. To overcome limitation \textbf{(2)}, we optimize the alignment loss solely through latent embeddings by stopping gradients on non-latent representations. Furthermore, we propose \textbf{V}iusal-\textbf{l}atent \textbf{P}olicy \textbf{O}ptimization (\textbf{VLPO}), which computes policy gradient directly for latent embeddings by estimating their output probability.

Our contributions are summarized as follows:
\begin{enumerate}
    \item We propose \textbf{Monet-SFT}, a three-stage supervised fine-tuning framework that trains MLLMs to generate and reason with latent embeddings.
    \item We propose \textbf{VLPO}, a novel RL algorithm tailored for latent reasoning. Unlike GRPO, which targets text reasoning, VLPO incorporates latent embeddings into the total loss by computing an approximate probability for the latent embeddings collected during rollout.
    \item We identify the limitations of existing image-text interleaved datasets: the unnecessary usage and inaccuracy of auxiliary images. To address these limitations, we further propose a multi-stage data curation pipeline to construct \textbf{Monet-SFT-152K}, a high-quality dataset with image-text interleaved chain-of-thoughts (CoT) for SFT.
    \item Extensive experiments on real-world perception and reasoning benchmarks show that Monet-SFT and VLPO outperform conventional SFT + GRPO, cropping-based methods, and prior latent visual reasoning approaches. VLPO further enhances out-of-distribution (OOD) generalization on unseen tasks.
\end{enumerate}

\section{Related Work}
\label{sec:related_work}

\subsection{Think with Images}
These studies can be broadly divided into two categories based on how they obtain intermediate-step auxiliary images:
(1) directly emphasizing visual content from the original image, such as grounding, cropping, or re-inputting selected image tokens \citep{sun2024visual, zheng2025deepeyes, zhang2025chain, sarch2025grounded, chung2025don, chen2025mint}; and
(2) creating new visual content beyond the original image, such as invoking external tools or code interpreters to edit the image (e.g., drawing lines, adding bounding boxes, computing depth maps) \citep{hu2024visual, zhao2025pyvision, su2025openthinkimg, chen2025learning, qi2024cogcom, fu2025refocus, zhang2025thyme, zhou2025reinforced}, or generating new images via text-to-image models \citep{chern2025thinking, li2025imagine, xu2025visual}.

While these approaches significantly enhance visual perception and reasoning, they also introduce key limitations.
First, models trained for specific visual tools, such as bounding box prediction, struggle to generalize to tasks requiring more complex visual operations (e.g., visual math, spatial, or graphic reasoning).
Second, tool-dependent reasoning increases the cognitive burden for training. Models often fail to generate valid tool calls or executable code without extensive supervision.
Finally, reliance on external tools or interpreters necessitates asynchronous, multi-turn inference, complicating deployment and increasing latency.

\subsection{Reasoning in Latent Space}
Recent studies have explored training LLMs to reason with continuous latent embeddings—either by replacing discrete text tokens with self-generated continuous embeddings \citep{hao2024training, geiping2025scaling, shen2025codi, wei2025simcotsupervisedimplicit, butt2025softtokenshard, wang2025synadapt} or by using reweighted combinations of text embeddings \citep{yang2025machine, pham2025multimodal}. This line of research aims to enhance reasoning flexibility by enabling exploration in a continuous latent space rather than a discrete linguistic space, while also shortening reasoning chains.

Building on this direction, several works have extended latent reasoning to MLLMs. \citet{yang2025machine, li2025latent} propose to align generated latent embeddings with those of auxiliary images. \citet{yang2025machine} further compresses image embeddings with mean pooling before alignment, which may potentially distort visual information, while \citet{li2025latent} focuses only on cropped image regions, which cannot encode visual operations over the entire image. \citet{pham2025multimodal} removes auxiliary images entirely, optimizing latent embeddings via next-token prediction, but with limited gains. Moreover, these methods neglect latent-space optimization during RL.
%Our approach addresses these limitations from several aspects. To improve the generality of latent visual embeddings, we construct Monet-SFT-152K, encompassing diverse visual operations (Table \ref{tab:sft_data_statistics}). Instead of aligning latent and image embeddings directly, we align the representations of key observation tokens that follow them and allow latent embeddings to attend to auxiliary image features, enabling the model to automatically encode essential visual information into latent embeddings.

\section{Method}
\label{sec:method}

\subsection{Overview}
\label{sec:prelimiaries}

\begin{figure*}
    \centering
    \includegraphics[width=1\linewidth]{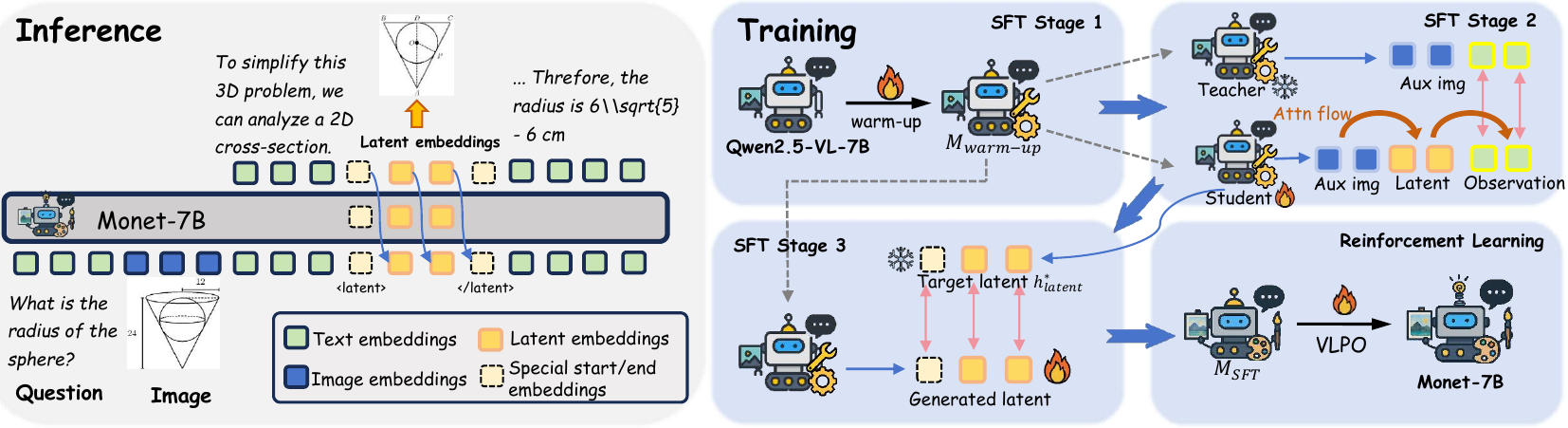}
    \caption{\textbf{Method overview.} \textit{Left}: During inference, Monet can automatically decide when to start latent reasoning by outputting a special start embedding. We fix the output length of the latent embeddings. \textit{Right}: We propose a three-stage SFT (Section \ref{sec:SFT}) and RL (Section \ref{sec:RL}) framework. The SFT stages progressively warm up the model, generate high-quality latent embeddings, and distill latent reasoning ability. The RL stage further refines the model using our VLPO algorithm, specifically designed for latent reasoning.}
    \label{fig:overview}
\vspace{-1.5em}
\end{figure*}
\textbf{Inference.} As illustrated in Figure \ref{fig:overview} (left), given a question and its corresponding image, Monet generates text-latent interleaved reasoning chains. Specifically, it can automatically decide when to generate a special token, ``\verb|<latent>|'', to initiate latent reasoning during inference. We modify the decoding process so that the representation from the last layer of the MLLM decoder is fed back as the next input embedding. After generating a predefined number $K$ of latent embeddings, we insert a special stop token, ``\verb|</latent>|'', prompting the model to switch back to language-based reasoning. Consistent with the findings of \citet{li2025latent}, this fixed-length decoding strategy proves simple yet effective. We provide a comprehensive analysis of the effect of $K$ in Section \ref{sec:latent_size}.

\textbf{Training.} The training of Monet includes a three-stage SFT and an RL stage. We briefly introduce the process and goal of each stage to provide an overview (Figure \ref{fig:overview} right).
\begin{enumerate}
    \item \textbf{SFT Stage 1.} Adapts the model to interleaved reasoning patterns. The warmed-up parameters $M_{\text{warm-up}}$ are used to initialize the teacher and student models in Stage 2.
    \item \textbf{SFT Stage 2.} Generates high-quality target latent embeddings $\mathbf{h^*}_{\text{latent}}$ for Stage 3. The student model learns to produce these embeddings by: (2) aligning its key observational token representations with those of the fixed teacher model; and (2) making latent embeddings directly attend to auxiliary image embeddings through a controlled attention flow.
    \item \textbf{SFT Stage 3.} Trains the model to generate latent embeddings without access to ground-truth auxiliary images. Initialized with $M_{\text{warm-up}}$, the model aligns the generated latent embeddings with the target embeddings $\mathbf{h^*}{\text{latent}}$ generated in Stage 2, resulting in the model $M_{\text{SFT}}$.
    \item \textbf{RL Stage.} Further optimizes $M_{\text{SFT}}$ using VLPO.
\end{enumerate}
In the following sections, we will elaborate on the design of each component.

\subsection{SFT Training Dataset Construction}
\label{sec:dataset}
\begin{figure*}[t]
    \centering
    \includegraphics[width=0.9\textwidth]{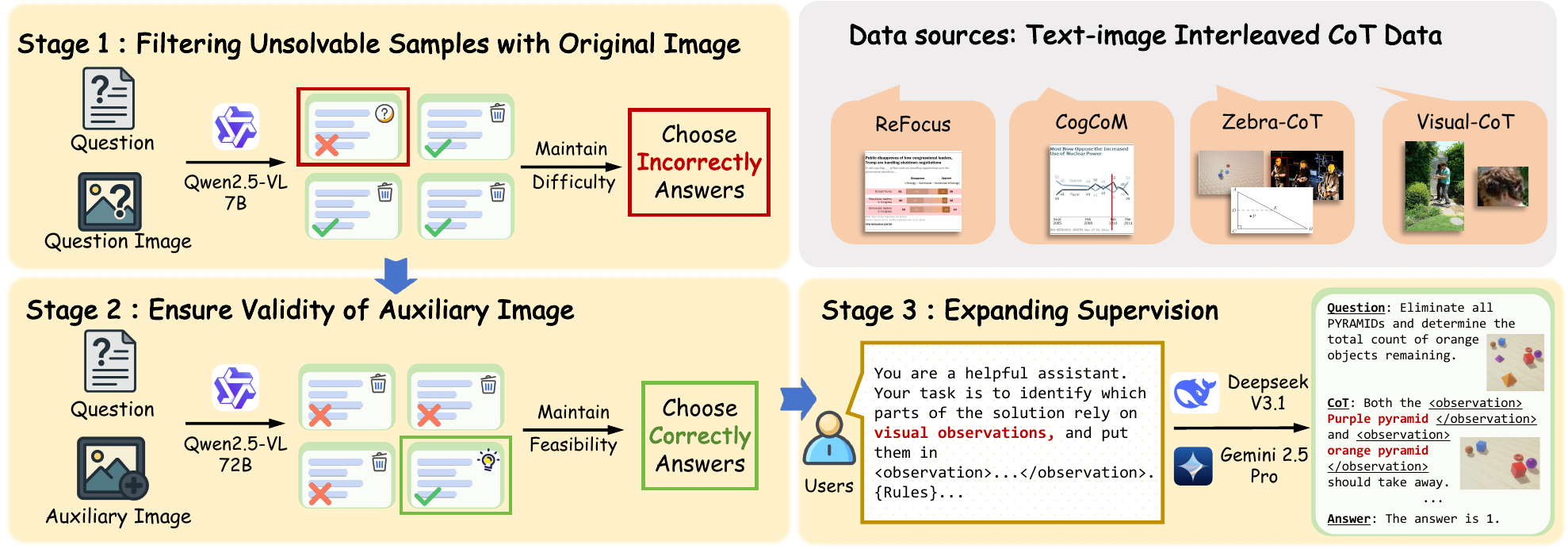}
    \caption{\textbf{Construction pipeline of Monet-SFT-125K}. Stage 1 filters hard samples (unsolvable from the original image). Stage 2 keeps those where auxiliary images lead to correct answers, ensuring their necessity and correctness. Stage 3 highlights key visual-observation tokens using advanced LLM judges, providing strong supervision for learning latent embeddings.}
    \label{fig:data_construction}
\end{figure*}

To train Monet, we require a dataset with image–text interleaved CoTs that provide supervision for intermediate latent thinking steps. However, existing datasets exhibit three major limitations:
(1) Many samples can be trivially solved by directly observing the input image, leading the model to bypass learning meaningful feature in the intermediate images.
(2) The intermediate images are sometimes inaccurate, introducing noise into training.
(3) All text tokens are treated equally, overlooking those that describe crucial visual information, which could serve as useful supervision signals \citep{fang2024wrong}.
To overcome these issues, we design a three-stage data curation pipeline that ensures the necessity and correctness of auxiliary images (Figure \ref{fig:data_construction}).

\textit{Stage 1}: We collect raw text-image interleaved CoT data from ReFocus \citep{fu2025refocus}, CogCoM \citep{qi2024cogcom}, Zebra-CoT \citep{li2025zebra}, and Visual-CoT \citep{shao2024visual}, and retain samples that Qwen2.5-VL-7B \cite{bai2025qwen2} \textbf{incorrectly} answers using only the question and the original image to ensure the necessity of auxiliary images. \textit{Stage 2}: From these samples, we keep those that Qwen2.5-VL-72B can \textbf{correctly} solve using only the auxiliary images, ensuring that the auxiliary images are accurate for reasoning. \textit{Stage 3}: We use Deepseek-V3.1 \citep{liu2024deepseek} and Gemini 2.5 Pro \citep{comanici2025gemini} to identify text tokens corresponding to crucial visual observations for answering the final question to provide fine-grained supervision for learning latent embeddings. These steps yield \textbf{Monet-SFT-125k}, in which the auxiliary images include cropping, grounding, highlighting, and operations that create new visual references for fine-grained perception and complex visual reasoning tasks. 
Table \ref{tab:sft_data_statistics} presents the statistics of Monet-SFT-125k.

\begin{table*}[htbp]
\caption{\textbf{Statistics of the Monet-SFT-125K dataset.} It contains 125K real-world, document, chart, and geometry problems with image–text interleaved CoTs, featuring visual operations ranging from simple extractive steps such as cropping and grounding to more complex ones, including drawing auxiliary lines and creating entirely new images that represent intermediate visual states.}
\label{tab:sft_data_statistics}
\centering
    \resizebox{0.9\textwidth}{!}{
        \begin{tabular}{l l l r}
        \hline
        \textbf{Data Source} & \textbf{Problem Domain} & \textbf{Visual Operation Type} & \textbf{Amount} \\
        \hline
        ReFocus \citep{fu2025refocus} &  Chart  & Drawing bounding boxes, highlighting  & 0.4K \\
        CogCoM \citep{qi2024cogcom}  & Real-world, chart  & Cropping, drawing auxiliary lines and bounding boxes          & 0.5K \\
        Visual-CoT \citep{shao2024visual} & Real-world, documents, chart & Cropping, drawing bounding boxes  & 118.6K \\
        Zebra-CoT visual search \citep{li2025zebra} & Real-world, documents, chart  & Cropping, drawing bounding boxes & 2.7K \\
        Zebra-CoT geometry \citep{li2025zebra}  &   Geometry     & Drawing auxiliary lines, creating new geometric sketchpads & 0.1 K \\
        Zebra-CoT count \citep{li2025zebra} &  3D object counting   & Creating new images with 3D objects removed or added      & 2.9K \\
        \hline
        \end{tabular}
}
\end{table*}

\begin{figure*}[t]
    \centering
    \includegraphics[width=0.95\textwidth]{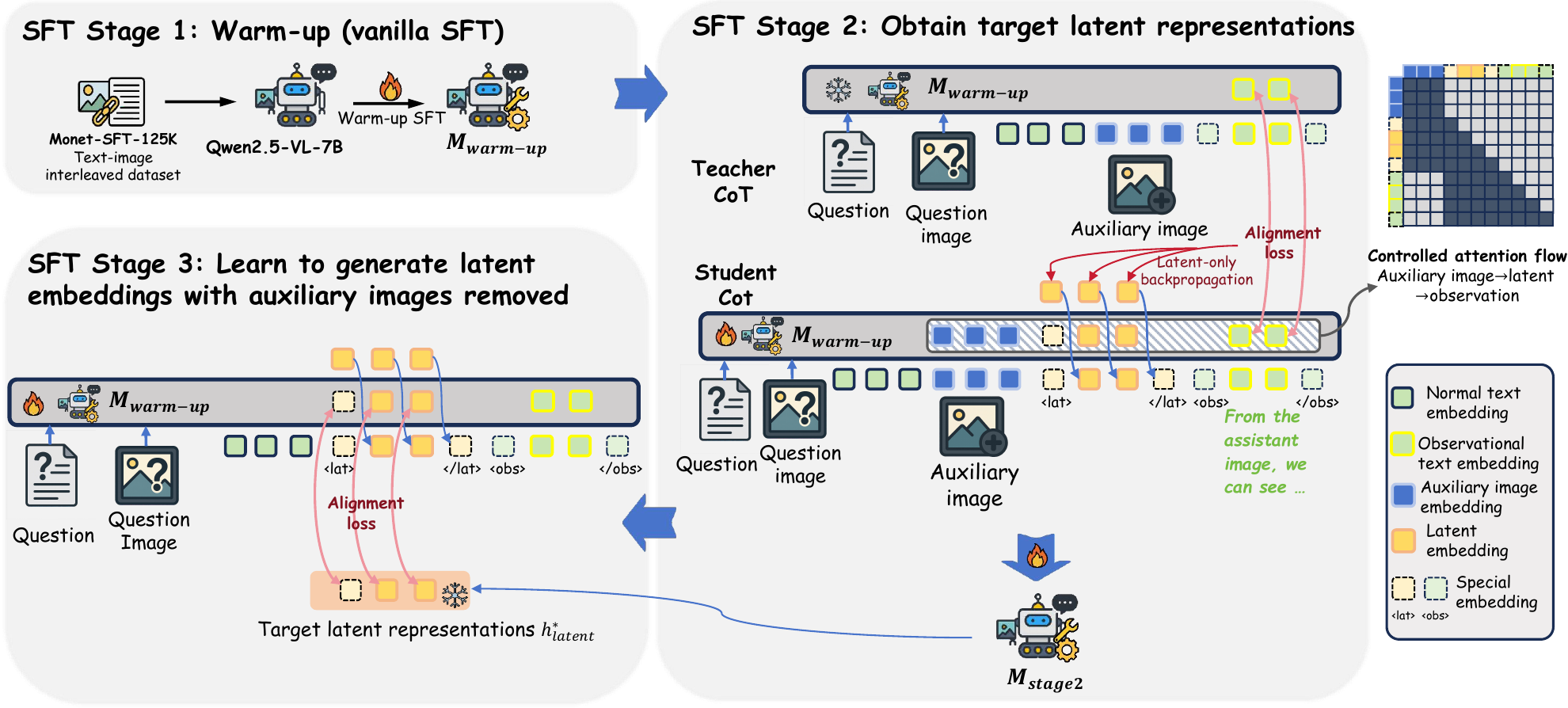}
    \vspace{-5pt}
    \caption{\textbf{The proposed three-stage SFT pipeline:} warm-up, supervised latent–observation alignment with controlled attention flow, and latent generation without auxiliary-image access.}
    \vspace{-15pt}
    \label{fig:method}
\end{figure*}

\subsection{Supervised Fine-tuning}
\label{sec:SFT}

The SFT stage addresses two key challenges in latent reasoning training of MLLMs:\textbf{ (1) Latent–visual alignment is computationally expensive,} as auxiliary images contain hundreds or thousands of image embeddings. \textbf{(2) The next-token-prediction objective provides weak supervision for latent embeddings.} During SFT, current methods \citep{pham2025multimodal, yang2025machine, li2025latent} employ cross-entropy loss on text tokens following the latent embeddings, allowing gradients to flow back to the latent embeddings. However, the model can simply memorize the following tokens instead of learning effective latent representations.

To overcome these issues, we introduce a three-stage SFT framework (Figure \ref{fig:method}) that enables the model to selectively encode useful visual features from auxiliary images while focusing on the downstream reasoning process, without incurring the cost of explicit latent–visual alignment. We now describe the design of each stage in detail.

\subsubsection{SFT Stage 1: Warm up.}
We first warm up the base model $M_{\text{base}}$ (Qwen2.5-VL-7B) by performing vanilla SFT on the image-text interleaved CoTs of Monet-SFT-125K, obtaining $M_{\text{warm-up}}$ (see Figure \ref{fig:method} upper left). This stage adapts the model to the image–text interleaved reasoning pattern so it can effectively leverage intermediate-step images when predicting subsequent tokens. Without this adaptation, the model tends to ignore auxiliary images, and the representations of observation tokens may fail to capture sufficient visual information.
\begin{figure}[t]
    \centering
    % \vspace{-10pt}
    \includegraphics[width=0.75\linewidth]{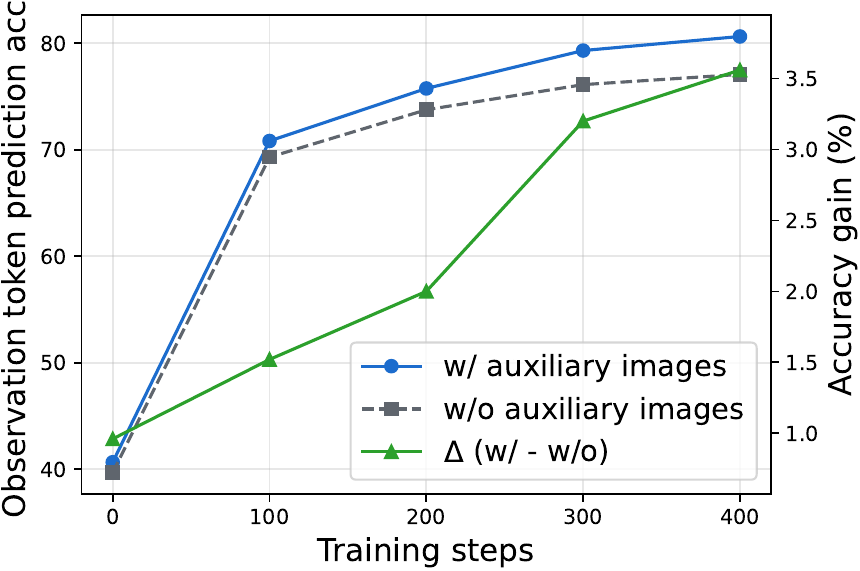}
    % \vspace{-10pt}
    \caption{\textbf{Prediction accuracy of the observation tokens during warm-up.} Training on image–text interleaved data encourages the model to utilize intermediate visual cues.}
    \label{fig:obs_acc}
    \vspace{-1em}
\end{figure}

Figure \ref{fig:obs_acc} illustrates this behavior, in which we compare the token prediction accuracy of the observation tokens with/without the auxiliary images during this warm-up stage: for the unadapted base model, using auxiliary images yields almost no improvement in predicting observation tokens, indicating poor utilization of intermediate visual cues. As the warm-up proceeds, however, the accuracy gain from auxiliary images gradually increases. This demonstrates that SFT on image–text interleaved data drives the model to rely more on intermediate-step visual features rather than memorizing language patterns. Consequently, observation tokens encode meaningful visual information from the auxiliary images and serve as effective supervision signals.

\subsubsection{SFT Stage 2: Obtain high-quality target latent embeddings.} In this stage, we train the model to generate latent embeddings that capture useful visual features from auxiliary images to support visual reasoning. We initialize both a teacher and a student model from $M_{\text{warm-up}}$. The teacher processes CoTs with ground-truth auxiliary images, while in the student CoT, each auxiliary image segment is followed by autoregressively-generated latent embeddings, and the auxiliary images are made visible only to these latent embeddings via a modified attention mask (Figure \ref{fig:method} right). Now we elaborate on our key designs as follows.

\textbf{Alignment on Key Observation Tokens.} Since latent embeddings are intended to serve the role of auxiliary images in predicting observation tokens, the hidden representations of these tokens should match those obtained when the ground-truth auxiliary images are provided. Inspired by recent text–based latent reasoning work \citep{shen2025codi, wang2025synadapt}, we therefore align the hidden representations of observation tokens under these two conditions.

Specifically, for each training sample, we freeze $M_{\text{warm-up}}$ and extract the observation-token representations from all layers when using auxiliary images, yielding $\mathcal{H}_{\text{obs}}^*=\{\mathbf{h^*}^{(i,l)}_{\text{obs}}\}_{i=1}^N$, where $i$ indexes samples and $l$ indexes layers, $N$ is the number of training samples. Let $\hat{\mathbf{h}}^{(i,l)}_{\text{obs}}$ denote the corresponding representations in the student CoT that uses generated latent embeddings. We fix $\mathbf{h^*}^{(i,l)}_{\text{obs}}$ and maximize their cosine similarity through the alignment loss:
\begin{equation}
    \mathcal{L}_{\text{align-obs}} = \frac{1}{N}\sum_i \sum_l \left(1-\cos ({\mathbf{h^*}^{(i, l)}_{\text{obs}}}.\operatorname{detach()}, ~{\hat{\mathbf{h}}^{(i, l)}_{\text{obs}}}) \right).
\end{equation}
 
To ensure that the observation alignment loss $\mathcal{L}_{\text{align-obs}}$ can only be minimized by optimizing the latent embeddings instead of through any undesired shortcuts, we restrict gradient flow from $\mathcal{L}_{\text{align-obs}}$ to pass solely through the generated latent embeddings to the model parameters. Implementation details are in the supplementary material. Empirically, removing this latent-only backpropagation leads to significant performance degradation (see Table \ref{tab:ablation_sft}, ``w/o latent-only BP'').

\textbf{``Auxiliary image $\rightarrow$ latent $\rightarrow$ observation'' Attention Flow.}
We observe that using the above alignment loss alone yields suboptimal performance (Table \ref{tab:ablation_sft}, “w/o auxiliary img”), as the observation-token representations may not encode sufficient visual information from the auxiliary images. To address this, for the student CoT, we insert the auxiliary image embeddings immediately before each latent-embedding segment and apply a modified attention mask that allows these image embeddings to be attended \textbf{only} by the latent embeddings, not by subsequent text tokens. This design offers two benefits:
first, latent embeddings can directly access the visual features from auxiliary images without information loss;
second, it enforces a structured flow of visual information, i.e., auxiliary images$\rightarrow$latent embeddings$\rightarrow$observation tokens, encouraging the latent embeddings to selectively encode the relevant visual cues.

\textbf{Next-Token Prediction Loss.} We Additionally apply a standard next-token prediction loss on text tokens to optimize the latent embeddings. Denote the text tokens in student CoT $i$ as $\mathcal{T}^{(i)}=\{\mathbf{y}_t\}_{t=1}^{|\mathcal{T}^{(i)}|}$. For each $\mathbf{y}_t$, we use $\mathbf{y}_{<t}$ to represent the context, which includes the question text, question image, previous text responses, and latent embeddings (note that the auxiliary images are invisible to the text tokens under our attention design). Denote the output probability distribution as $p$, the next-token-prediction loss is:
\begin{equation}
\label{loss:ntp}
    \mathcal{L}_{\text{NTP}} = -\frac{1}{N} \sum_i \sum_{\mathbf{y}_t\in \mathcal{T}^{(i)}} \log p (\mathbf{y}_t \mid \mathbf{y}_{<t} ).
\end{equation}

\textbf{Total Loss of SFT Stage 2.} The total loss is:
\begin{equation}
    \mathcal{L}_{\text{stage2}} = \mathcal{L}_{\text{NTP}} + \alpha \mathcal{L}_{\text{align-obs}},
\end{equation}
in which we set $\alpha=2.0$ in all experiments. We denote the student model after training as $M_{\text{stage2}}$. After this stage, we use $M_{\text{stage2}}$ to generate the latent embeddings, denoted as $\mathbf{h^*}^{(i)}_{\text{latent}}$, which serve as target latent embeddings used for the next stage.

\begin{table}[t]
    \centering
    \caption{\textbf{Ablation of the components of the SFT Stage 2.} ``Monet-SFT'' denotes the full SFT pipeline. ``w/o latent-only BP'' represents without latent-only backpropagation for the alignment loss. ``w/o auxiliary img'' disables attention from latent embeddings to auxiliary-image embeddings in the student CoT. Both ablation lead to performance degradation on almost all tasks.}
    \label{tab:ablation_sft}
    \resizebox{0.48\textwidth}{!}{%
    \begin{tabular}{lccccc}
        \toprule
        & \textbf{V*} & \textbf{HR4K} & \textbf{HR8K} & \textbf{MME-RealWorld} & \textbf{VisualPuzzles}\\
        \midrule
        \rowcolor{gray!20}
        Monet-SFT & 82.20 & 68.50 & 66.00 & 55.03 & 30.48 \\
        \quad w/o latent-only BP & 46.07 & 40.13 & 39.00 & 38.67 &  33.65\\
        \quad w/o auxiliary img & 73.30 & 63.88 & 57.63 & 39.66 & 28.60 \\
        \bottomrule
    \end{tabular}%
    }
\end{table}

\subsubsection{SFT Stage 3: Learn to generate latent embeddings without auxiliary images.} After Stage 2, we obtain target latent embeddings $\mathbf{h}^{*(i)}_{\text{latent}}$ that encode the information contained in auxiliary images. However, the model $M_{\text{stage2}}$ is trained in a setting where latent embeddings can still attend to auxiliary images, which is different from our ultimate goal of generating latent embeddings without ground-truth auxiliary images. To close this gap, we reinitialize the model with $M_{\text{warm-up}}$ and train it to produce latent embeddings $\hat{\mathbf{h}}^{(i)}_{\text{latent}}$ with auxiliary images removed in the CoT (Figure \ref{fig:method} bottom left). We then align them with the fixed target embeddings $\mathbf{h}^{*(i)}_{\text{latent}}$ using the following loss:

\begin{equation}
    \mathcal{L}_{\text{align-latent}} = \frac{1}{N}\sum_i \sum_l \left( 1-\cos ({\mathbf{h^*}^{(i, l)}_{\text{latent}}}.\operatorname{detach()}, ~{\hat{\mathbf{h}}^{(i, l)}_{\text{latent}}}) \right).
\end{equation}
Unlike previous latent-visual reasoning work \citep{yang2025machine, li2025latent}, which aligns only the final-layer representations, we align all layers to provide stronger supervision.

We also apply a next-token-prediction loss $\mathcal{L}_{NTP}$ on the text tokens to let the latent embeddings benefit subsequent reasoning, which has the same form as Equation (\ref{loss:ntp}).

\textbf{Total loss of SFT Stage 3.} The total loss of Stage 3 is:
\begin{equation}
    \mathcal{L}_{\text{stage3}} = \mathcal{L}_{\text{NTP}} + \beta \mathcal{L}_{\text{align-latent}}.
\end{equation}
in which we set $\beta=2.0$ in all our experiments.

\subsection{VLPO: Visual-latent Policy Optimization}
\label{sec:RL}
\textbf{Limitations of GRPO on latent visual reasoning.} Prior latent visual reasoning works \citep{yang2025machine, li2025latent} apply GRPO \citep{Shao2024DeepSeekMathPT} after SFT. Given a question $Q$ and a question image $I$, the old policy model $\pi_{\text{old}}$ generates a group of responses $\{\mathbf{o}_1, \mathbf{o}_2, ..., \mathbf{o}_G\}$. Then GRPO updates the policy model $\pi_{\theta}$ by maximizing:
\begin{equation}
    \begin{aligned}
        &\mathcal{J}_{GRPO}(\theta)=\mathbb{E}_{Q, I, o\sim \pi_{\text{old}}}\\
        &\frac{1}{G} \sum_{i=1}^G \frac{1}{\left|\mathbf{o}_i\right|} \sum_{t=1}^{\left|\mathbf{o}_i\right|}\min \left[ r_{i,t}(\theta) \hat{A}_{i, t}, \operatorname{clip}\left(r_{i,t}(\theta), 1-\varepsilon, 1+\varepsilon\right) \hat{A}_{i, t}\right]\\
        &-\beta \operatorname{K L}\left(\pi_\theta \| \pi_{\text{ref}}\right),
    \end{aligned}
\end{equation}

where $r_{i,t}(\theta)=\frac{\pi_\theta\left(\mathbf{o}_{i, t} \mid Q, I, \mathbf{o}_{i,<t}\right)}{\pi_{\theta_{\text {old }}}\left(\mathbf{o}_{i, t} \mid Q, I, \mathbf{o}_{i,<t}\right)}$, $\mathbf{o}_{i,<t}$ are the response tokens before position $t$, $\hat{A}_{i, t}$ is the advantage computed based on the outcome rewards $\{r_1,r_2,...,r_G\}$ of the responses, i.e., $\hat{A}_{i, t}=\frac{r_i-\text{mean}(\{r_1,r_2,...,r_G\})}{\text{std}(\{r_1,r_2,...,r_G\})}$. $\pi_{\text{ref}}$ is the reference model. A critical limitation is that the GRPO objective \textbf{can only be computed on text tokens}. Since latent embeddings have no explicit probability distribution like text tokens, GRPO cannot directly optimize them, leaving the latent reasoning component largely untrained during RL.

To address this limitation, we propose Visual–latent Policy Optimization (VLPO), a novel reinforcement learning objective tailored for latent reasoning. The key idea of VLPO is to estimate the \textbf{output probability of the continuous latent embeddings} collected during rollout to compute $r_{i,t}(\theta)$ for the latent embeddings. This enables latent embeddings to be optimized directly with outcome rewards, just like text tokens. Denote the latent embedding generated by $\pi_{\text{old}}$ at position $t$ of the $i$-th rollout as $\mathbf{h}^{\text{old}}_{i,t}$ with context $[Q,I,\mathbf{o}_{i,<t}]$. To compute $\pi_{\theta}(\mathbf{h}^{\text{old}}_{i,t}\mid Q,I,\mathbf{o}_{i,<t})$, we model $\mathbf{h}^{\text{old}}_{i,t}$ as a sample drawn from a latent Gaussian distribution whose mean is the latent embedding generated by $\pi_{\theta}$ given the same context. We denote this mean by $\mathbf{h}^{\theta}_{i,t}$. Under this assumption, $\pi_{\theta}(\mathbf{h}^{\text{old}}_{i,t} \mid Q,I,\mathbf{o}_{i,<t})$ can be computed as:
\begin{equation}
    \begin{aligned}
        &\pi_{\theta}(\mathbf{h}^{\text{old}}_{i,t}\mid Q,I,\mathbf{o}_{i,<t}) = \exp\left( -\frac{1}{2\sigma^2}\|\mathbf{h}^{\text{old}}_{i,t}-\mathbf{h}^{\theta}_{i,t}\|^2 - const \right)
    \end{aligned}
\end{equation}
where $\sigma$ is a predefined scalar hyperparameter and $const$ is a constant. $\pi_{\theta}(o^{\text{latent}}_{i,t} \mid Q,I,\mathbf{o}_{i,<t})$ measures the probability that $\pi_{\theta}$ chooses $o^{\text{latent}}_{i,t}$ as the ``action'' at position $t$ of response $i$, enabling us to compute $r_{i,t}(\theta)$ for a latent step:
\begin{equation}
\label{eq:vlpo}
\begin{aligned}
    r_{i,t}(\theta) = &  \frac{\pi_\theta\left(\mathbf{h}^{\text{old}}_{i,t} \mid Q, I, \mathbf{o}_{i,<t}\right)}{\pi_{\theta_{\text {old }}}\left(\mathbf{h}^{\theta}_{i,t} \mid Q, I, \mathbf{o}_{i,<t}\right)}=\exp \left(-\frac{1}{2\sigma^2}\|\mathbf{h}^{\text{old}}_{i,t}-\mathbf{h}^{\theta}_{i,t}\|^2 \right)
\end{aligned}
\end{equation}
Replacing the original $r_{i,t}(\theta)$ for latent steps yields our VLPO objective. During optimization, we fix the rollout rollout $\mathbf{h}^{\text{old}}_{i,t}$ and optimize $\mathbf{h}^{\theta}_{i,t}$ generated by the policy.

\textbf{Reward Design.} We use an accuracy reward (1 for a correct answer; 0 otherwise) and a format reward encouraging the model to place the final answer in ``\verb|\boxed{}|''. Importantly, we do not reward latent-reasoning behavior itself, as doing so would incentivize the model to invoke latent reasoning indiscriminately. We find this simple reward scheme yields strong empirical performance.

\textbf{How does VLPO work?} When $\hat{A}_{i, t}>0$, maximizing the VLPO objective will increase $r_{i,t}(\theta)$. For text tokens, this means maximizing its output probability; For latent embeddings, Equation (\ref{eq:vlpo}) shows that maximizing this $r_{i,t}(\theta)$ equals to minimizing $\|\mathbf{h}^{\text{old}}_{i,t}-\mathbf{h}^{\theta}_{i,t}\|^2$, thereby \textbf{pulling the policy latent embeddings to a ``good-action'' latent embedding} $\mathbf{h}^{\theta}_{i,t}$ \textbf{that led to a positive outcome reward.} Thus, VLPO directly optimizes latent embeddings using reward signals—an ability that GRPO fundamentally lacks.
\begin{table*}[t]
    \centering
    \caption{\textbf{Performance on real-world perception and reasoning benchmarks.} The best-performing open-source model for each dataset is highlighted in \textbf{bold}. Results marked with ``*'' are reported by other papers \citep{zhang2025thyme, li2025latent}, while the others are from our own evaluations.}
    \label{tab:exp-id}
    \resizebox{\textwidth}{!}{%
    \begin{tabular}{lcccccccccccc}
        \toprule
        \multirow{2}{*}{\textbf{Model}} & \multicolumn{3}{c}{\textbf{V*}} & \multicolumn{3}{c}{\textbf{HRBench4K}} & \multicolumn{3}{c}{\textbf{HRBench8K}} & \multicolumn{3}{c}{\textbf{MME-RealWorld-Lite}} \\
        \cmidrule(lr){2-4}\cmidrule(lr){5-7}\cmidrule(lr){8-10}\cmidrule(lr){11-13}
        & Overall & Attribute & Spatial & Overall & FSP & FCP & Overall & FSP & FCP & Overall & Reasoning & Perception \\
        \midrule \rowcolor{red4!20}
        \multicolumn{13}{c}{\textit{\textbf{Proprietary Model}}}
        \\
        %Gemini 2.5 Pro~\cite{comanici2025gemini} &  &  &  &  &  &  &  &  &  &  &  &  \\
        GPT-4o~\cite{gpt_4o} & 67.5* & 72.2* & 60.5* & 59.0* & 70.0* & 48.0* & 55.5* & 62.0* & 49.0* & 52.0* & 48.3* & 54.4* \\
        \midrule \rowcolor{blue4!20}
        \multicolumn{13}{c}{\textit{\textbf{Open-Source Model}}}
        \\
        Qwen2.5-VL-7B~\cite{bai2025qwen2} & 76.44 & 77.39 & 75.00 & 68.00 & 80.25 & 55.75 & 63.75 & 73.75 & 53.75 & 45.75 & 39.73 & 49.62 \\
        \quad + vanilla SFT & 81.68 & \textbf{83.48} & 78.95 & 68.38 & 78.28 & 58.5 & 61.63 & 70.75 & 52.50 & 51.28 & 46.4 & 54.41 \\
        \quad + vanilla SFT + GRPO & 78.53 & 78.26 & 78.95 & 70.00 & 83.25 & 56.75 & 66.75 & 78.00 & 55.50 & 52.42 & 48.13 & 55.18 \\
        Deepeyes \citep{zheng2025deepeyes} & \textbf{83.25} & \textbf{84.35} & 81.58 & \textbf{71.25} & 83.75 & \textbf{58.75} & 65.13 & 77.00 & 53.25 & 54.28 & 50.53 & 56.63 \\
        LVR \citep{li2025latent} & 80.6* & 81.7* & 79.0* & - & - & - & - & - & - & - & - & - \\
        % \rowcolor{gray!20}
        \midrule \rowcolor[HTML]{F5FFFA}
        \multicolumn{13}{c}{\textit{\textbf{Our Model}}}
        \\
        \textbf{Monet-7B (SFT + VLPO)} & \textbf{83.25} & 83.48 & \textbf{82.89} & 71.00 & \textbf{85.25} & 56.75 & \textbf{68.00} & \textbf{79.75} & \textbf{56.25} & \textbf{55.50} & \textbf{51.07} & \textbf{58.34} \\
        % \rowcolor{gray!20}
        \textit{Relative Improvement} & \textcolor[rgb]{0,0.6,0}{+6.81} & \textcolor[rgb]{0,0.6,0}{+6.09} & \textcolor[rgb]{0,0.6,0}{+7.89} & \textcolor[rgb]{0,0.6,0}{+3.00} & \textcolor[rgb]{0,0.6,0}{+5.00} & \textcolor[rgb]{0,0.6,0}{+1.00} & \textcolor[rgb]{0,0.6,0}{+4.25} & \textcolor[rgb]{0,0.6,0}{+6.00} & \textcolor[rgb]{0,0.6,0}{+2.50} & \textcolor[rgb]{0,0.6,0}{+9.75} & \textcolor[rgb]{0,0.6,0}{+11.34} & \textcolor[rgb]{0,0.6,0}{+8.72} \\
        \bottomrule
    \end{tabular}%
    }
\end{table*}

\section{Experiment}
\label{sec:experiment}

\subsection{Experiment Setup}
\textbf{Training and Evaluation Setup.} In SFT Stage 1, we train the model for 4 epochs. For SFT stage 2 and 3, we train 1000 steps (about 1 epoch). For RL training, we adopt a 3.2K subset of Thyme-RL \citep{zhang2025thyme} and train for 1 epoch. For Monet-SFT model, we fix the training time latent size to 8; For \textbf{Monet-7B} (Monet-SFT + VLPO), we use a training latent size of 10. At test time, we select the best latent size from $\{8, 10, 12, 16\}$. The complete training and experimental setups are provided in the supplementary material.%\ref{app:implementation_details} and \ref{app:detailed_experimental_setup}, respectively. 

\textbf{Evaluated Benchmarks.} We adopt the VLMEvalKit \citep{duan2024vlmevalkit} framework for fair evaluation. We evaluate Monet on V* \citep{wu2024v}, HRBench4k~\citep{wang2025divide}, HRBench8K \citep{wang2025divide}, MME-RealWorld \citep{zhang2024mme}, covering fine-grained perception and reasoning tasks on real-world, chart and OCR data. To evaluate OOD capabilities, we also evaluate on VisualPuzzles \citep{song2025visualpuzzles}, which includes visual logical puzzles that demands abstract visual reasoning abilities rather than prior knowledge.

\textbf{Baselines.} We compare Monet against the following baselines: (1) \textbf{Qwen2.5-VL-7B}, the base model to train Monet-7B; (2) \textbf{Vanilla SFT}, which directly train Qwen2.5-VL-7B with Monet-SFT-125K; (3) \textbf{Vanilla SFT + GRPO}, which applies GRPO after Vanilla SFT; (4) \textbf{Deepeyes} \citep{zheng2025deepeyes}, a representative ``think with images'' approach that performs cropping to improve perception; (5) \textbf{LVR} \citep{li2025latent}, a recent work on latent visual reasoning that aligns the latent and auxiliary image embeddings; (6) other high-performance MLLMs, including Gemini 2.5 Pro \citep{comanici2025gemini}, GPT-4o \citep{gpt_4o}, Pangea-7B \citep{yue2024pangea}, and LLaVA-OneVision-72B \citep{li2024llava}.

\subsection{Main Results}
From Table \ref{tab:exp-id} and Tabel \ref{tab:exp-ood}, we can see that: (1) \textbf{Monet substantially enhances both perception and reasoning on real-world, chart, and OCR tasks, consistently outperforming the baselines.} It improves Qwen2.5-VL-7B by 4.25\%–9.75\%, surpasses both Vanilla SFT and SFT + GRPO trained on the same data, and outperforms Deepeyes and LVR on most benchmarks. (2) \textbf{Monet demonstrates strong OOD generalization.} It achieves the best performance on VisualPuzzles, which mainly contains abstract visual reasoning problems that are unseen during training.
% Requires: \usepackage{multirow}

\begin{table}[t]
    \centering
    \caption{\textbf{Performance on the VisualPuzzles benchmark (out-of-distribution).} The best-performing open-source model for each dataset is highlighted in \textbf{bold}. Results marked with ``*'' are taken from the original VisualPuzzles~\cite{song2025visualpuzzles} paper, while the others are obtained from our own evaluations.}
    \label{tab:exp-ood}
    \resizebox{0.48\textwidth}{!}{%
    \begin{tabular}{lcccccc}
        \toprule 
        \multirow{2}{*}{\textbf{Model}} & \multicolumn{6}{c}{\textbf{VisualPuzzles}} \\
        \cmidrule(lr){2-7}
        & Overall & Algorithmic & Analogical & Deductive & Inductive & Spatial  \\
        \midrule \rowcolor{red4!20}
        \multicolumn{7}{c}{\textit{\textbf{Proprietary Models}}}
        \\
        Gemini 2.5 Pro \citep{comanici2025gemini} & 49.5* & 60.0* & 64.0* & 60.0* & 29.7* & 36.4* \\
        GPT-4o \citep{gpt_4o} & 41.3* & 49.2* & 58.3* & 49.0* & 27.3* & 26.2* \\
        \midrule \rowcolor{blue4!20}
        \multicolumn{7}{c}{\textit{\textbf{Open-Source Models}}}
        \\
        Qwen2.5-VL-7B & 32.71 & 37.02 & 21.80 & 47.50 & 26.32 & 21.80  \\
        \quad + vanilla SFT & 33.99 & 40.46 & \textbf{30.81} & 46.00 & 24.40 & \textbf{29.02}  \\
        \quad + vanilla SFT + GRPO & 30.99 & 36.26 & 25.12 & 43.50 & 23.45 & 27.27 \\
        Deepeyes \citep{zheng2025deepeyes} & 32.96 & 37.79 & 27.01 & 41.00 & 26.79 & 27.01 \\
        Pangea-7B \citep{yue2024pangea} & 31.3* & 32.4* & 23.7* & 38.5* & \textbf{28.7}* & 32.5* \\
        LLaVA-OneVision-72B \cite{li2024llava} & 30.8* & 34.7* & 26.5* & 37.0* & 27.3* & 28.7* \\
        \midrule \rowcolor[HTML]{F5FFFA}
        \multicolumn{7}{c}{\textit{\textbf{Our Model}}}
        \\
        \textbf{Monet-7B} & \textbf{35.02} & \textbf{45.80} & \textbf{30.81} & \textbf{47.50} & 26.79 & 25.52  \\
        \textit{Relative Improvement} & \textcolor[rgb]{0,0.6,0}{+2.31} & \textcolor[rgb]{0,0.6,0}{+6.09} & \textcolor[rgb]{0,0.6,0}{+8.78} & \textcolor[rgb]{0,0.6,0}{+9.01} & \textcolor[rgb]{0,0.6,0}{+0.47} & \textcolor[rgb]{0,0.6,0}{+3.72} \\
        \bottomrule
    \end{tabular}%
    }
\end{table}

\subsection{Analysis of the Role of the Components Design}
\label{sec:ablation}
In this section, we present ablation studies to assess the necessity and effectiveness of each component in our training framework (Table \ref{tab:ablation}). We also discuss several unsuccessful attempts from our early exploration of training objectives and pipeline designs, which we hope can provide insights for future advancements in latent reasoning for MLLMs.

\begin{figure*}
    \centering
    \includegraphics[width=1\linewidth]{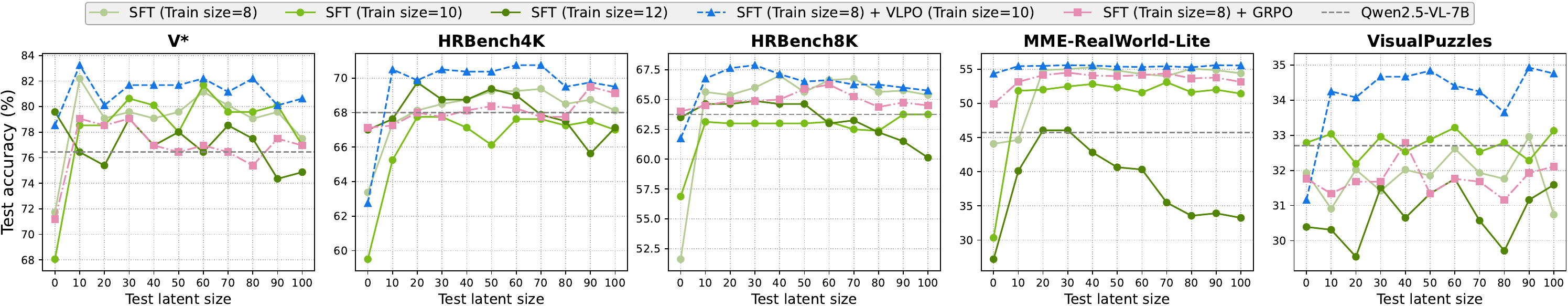}
    \caption{\textbf{Effect of the number of abstract visual embeddings used during training and inference on test accuracy.} The dashed line marks the accuracy of Qwen2.5-VL-7B. }
    \label{fig:latent_size}
    \vspace{-15pt}
\end{figure*}

\begin{table}[t]
    \centering
    \caption{\textbf{Ablation of the components of Monet.}}
    \label{tab:ablation}
    \resizebox{0.48\textwidth}{!}{%
    \begin{tabular}{lccccc}
        \toprule
        \textbf{Model} & \textbf{V*} & \textbf{HRBench4K} & \textbf{HRBench8K} & \textbf{MME-RealWorld-Lite} & \textbf{VisualPuzzles}\\
        \midrule
        \rowcolor{gray!20}
        \textbf{Monet-7B (full)} & \textbf{83.25} & \textbf{71.00} & \textbf{68.00} & \textbf{55.50} & \textbf{35.02} \\
        Monet-SFT (w/o VLPO) & 82.20 & 68.50 & 66.00 & 52.68 & 30.48 \\
        \quad + GRPO & 80.10 & 69.00 & 64.75 & 54.19 & 31.51 \\
        \quad w/o latent-only BP & 46.07 & 40.13 & 39.00 & 38.67 &  33.65\\
        \quad w/o auxiliary img & 73.30 & 63.88 & 57.63 & 39.66 & 28.60 \\
        \quad w/o obs token align & 75.39 & 67.25 & 63.50 & 46.90 & 27.48 \\
        \bottomrule
    \end{tabular}%
    }
\end{table}

\textbf{Takeaway 1. Single-signal supervision is less effective.} We remove the representation-alignment loss on observation tokens in SFT Stage 2 while retaining auxiliary-image embeddings in the student CoT under the ``image$\rightarrow$latent$\rightarrow$observation'' attention flow, which reflects our initial attempt. As shown by ``A-SFT w/o obs. token align'', removing the alignment loss causes a substantial drop in performance. This indicates that supervision from observation tokens is crucial. Similarly, removing the auxiliary images visible to latent embeddings while using the alignment loss alone (``w/o auxiliary img'') also significantly degrades performance, showing the importance of visual supervision on the latent embeddings. These two ablations demonstrate the necessity of the dual supervisions we use.

\textbf{Takeaway 2. GRPO doesn't consistently improve Monet-SFT, whereas VLPO does.} This is evident when comparing ``Monet-SFT'' with ``Monet-SFT + GRPO'', indicating that GRPO is not well-suited for latent reasoning. In contrast, applying VLPO significantly enhances the performance of Monet-SFT.

\textbf{Takeaway 3. Latent-related losses must backpropagate only through the latent embeddings.} ``w/o latent-only BP'' shows that when the alignment loss is allowed to update non-latent representations, performance drops sharply. This indicates that the model may otherwise exploit shortcut paths by minimizing the loss without actually improving the latent embeddings. Therefore, enforcing latent-only backpropagation is necessary.

\subsection{Effect of the Number of the Latent Embeddings on Performance}
\label{sec:latent_size}
In this section, we analyze how training and test-time latent size ($K_{\text{train}}$ and $K_{\text{test}}$) affect performance. For Monet-SFT, we choose $K_{\text{train}}\in\{8, 10, 12\}$. For RL models, we take Monet-SFT with $K_{\text{train}}=8$ and further apply VLPO (with RL $K_{\text{train}}=10$) and GRPO. Results are shown in Figure \ref{fig:latent_size}.

\textbf{Takeaway 1. For in-distribution tasks, improvements indeed come from using abstract visual embeddings; while for OOD tasks, only VLPO incentivizes useful latent reasoning.} Across all in-distribution tasks (V*, HRBench, MME-RealWorld), models perform worst when $K_{\text{test}}=0$. However, on the OOD task (VisualPuzzles), only the VLPO-enhanced model consistently benefits from latent reasoning, i.e., using a $K_{\text{test}}>0$ induces better performances than $K_{\text{test}}=0$, indicating that SFT alone cannot induce strong OOD generalization.

\textbf{Takeaway 2. For in-distribution tasks, Monet-SFT enables test-time scaling of latent embeddings; VLPO further extends this trend to OOD scenarios.} On V*, HRBench, and MME-RealWorld, performance typically peaks at a test-time latent size $K_{\text{test}}$ larger than the training-time one $K_{\text{train}}$. Before the peak, the performance improves as $K_{\text{test}}$ increases.

\textbf{Takeaway 3. VLPO improves robustness to the choice of $K_{\text{test}}$.} Compared with other models, Monet-SFT + VLPO exhibits the most stable performance as $K_{\text{test}}$ varies.

\textbf{Takeaway 4. GRPO mainly strengthens non-latent reasoning and provides limited benefits for latent reasoning.} When $K_{\text{test}}=0$, Monet-SFT ($K_{\text{train}}=8$) + GRPO is comparable to or better than Monet-SFT ($K_{\text{train}}=8$), showing gains in non-latent reasoning; However, when $K_{\text{test}}>0$, Monet-SFT + GRPO underperforms the SFT model for most $K_{\text{test}}$, suggesting minimal improvements to latent reasoning. This aligns with our analysis in Section \ref{sec:RL}: GRPO optimizes only text tokens and thus provides no direct supervision for latent embeddings.
\section{Conclusion and Limitations}
\label{sec:conclusion}
We introduce Monet, a framework that enables MLLMs to perform visual reasoning in the latent space. Our Monet-SFT pipeline enables the model to selectively encode useful visual cues from auxiliary images into compact latent embeddings, while our VLPO algorithm explicitly optimizes these embeddings during RL. We further construct the Monet-SFT-125K dataset to support SFT. Experiments show that Monet substantially improves real-world perception and reasoning performance and exhibits strong OOD generalization on abstract visual reasoning tasks. Together, these results demonstrate Monet as a promising direction toward more flexible and general multimodal reasoning.

We summarize the limitations of our work as follows. First, Monet relies on a multi-stage SFT pipeline, which may increase the overall training complexity and overhead. Second, we have not yet explored how different reward designs might influence latent visual reasoning in MLLMs, leaving room for exploration and further enhancement.

{
    \small
    \bibliographystyle{ieeenat_fullname}
    \bibliography{main}
}

\clearpage
\setcounter{page}{1}
\setcounter{section}{0} % 重置章节计数器
\renewcommand{\thesection}{\Alph{section}} % 将章节编号改为字母形式

\maketitlesupplementary

\section{Additional Experimental Results}
\label{app:additional_experiment_results}
\begin{table*}[htbp]
    \centering
    \caption{\textbf{Full result on MME-RealWorld-Lite.}}
    \label{tab:mme_full}
    \resizebox{0.9\textwidth}{!}{%
    \begin{tabular}{lccccc|cccccc}
        \hline
        & \multicolumn{11}{c}{\textbf{MME-RealWorld-Lite}} \\
        \cmidrule{2-12}
        & \multicolumn{5}{c}{\textbf{Reasoning}} & \multicolumn{6}{c}{\textbf{Perception}} \\
        \cmidrule{2-6} \cmidrule{7-12}
        & Overall & MO & AD & OCR & DT & overall & MO & AD & OCR & DT & RS  \\
        \hline
        %GPT-4o & 37.61 & 36.51 & 26.41 & 61.40 & 44.80   & 43.44 & 25.20 & 21.14 & 55.90 & 54.86 & 23.67 \\
        Qwen2.5-VL-7B-Instruct & 39.73 & 30.67 & 27.00  & 74.00 & 70.00 & 49.62 & 31.03 & 31.43 & 90.80 & 85.00 & 39.33 \\
        \quad + SFT & 46.40 & 52.00 & 33.00 & 73.00 & 65.00 & 54.41 & 40.44 & 36.29 & 88.80  & \textbf{86.00} & 48.00 \\
        \quad + SFT + GRPO & 48.13 & \textbf{59.33} & 33.75 & 71.00 & 66.00 & 55.18 & 43.26  & 37.71 & 89.60 & 83.00 & 45.33 \\
        Deepeyes & 50.53 & 46.67 & 40.25 & \textbf{78.00} & 70.00 & 56.63 & \textbf{43.89} & 38.86 & \textbf{90.00} & 84.00 & 51.33 \\
        Monet-SFT & 50.53 & 48.67 & \textbf{43.00} & 65.00 & 69.00 & 54.06 & 41.69 & 44.57 & 76.40 & 71.00 & 54.00 \\
        Monet-7B (Monet-SFT + VLPO) & \textbf{51.07} & 46.00 & 41.50 & 73.00 & \textbf{75.00} & \textbf{58.34} & 41.07 & \textbf{48.86} & 85.60 & 84.00 & \textbf{54.67} \\
        \hline
    \end{tabular}
    }
\end{table*}
We present the complete results on MME-RealWorld-Lite in Table \ref{tab:mme_full}. This benchmark spans a wide range of task domains, such as monitoring, autonomous driving, and complex diagram data. Compared with baseline approaches, Monet achieves consistently strong performance across all data sources, demonstrating its generality in both visual reasoning and perception tasks.

\section{Implementation Details}
\label{app:implementation_details}
\subsection{SFT Training}
\label{app:sft}
We use TRL \citep{vonwerra2022trl} to implement our SFT training. In a training CoT, we allow the model to predict the special latent-start token \verb|<latent>| while masking the latent-end token \verb|</latent>|, the observation-start token \verb|<observation>|, and the observation-end token \verb|</observation>| in the SFT labels.

\textbf{Hyperparameters.}  We list the training hyperparameters of SFT in Table \ref{tab:hyperparameters_sft}. To avoid OOM, we use a batch size of 1 and set gradient accumulation steps to 16. Additionally, we restrict the total number of pixels (denoted by ``max total pixels'') in a training sequence to prevent the OOM issue caused by too many auxiliary images.
\begin{table}[htbp]
    \centering
    \caption{\textbf{Hyperparameters for SFT.} }
    \label{tab:hyperparameters_sft}
    \resizebox{0.45\textwidth}{!}{%
    \begin{tabular}{lc}
        \toprule 
        Hyperparameter & Value \\
        \midrule 
        learning rate & 0.00001 \\
        batch size & 1 \\
        gradient accumulation steps & 16 \\
        weight decay & 0.01 \\
        SFT Stage 1 steps & 3885 (3 epochs) \\
        SFT Stage 2 steps & 1000 (1.03 epoch) \\
        SFT Stage 3 steps & 1000 (1.03 epoch) \\
        latent size & $\{8,10,12\}$ \\
        SFT Stage 1 max total pixels & 2000$\times$28$\times$28 \\
        SFT Stage 2 max total pixels & 1500$\times$28$\times$28 \\
        SFT Stage 3 max total pixels & 2000$\times$28$\times$28 \\
        SFT Stage 1,2,3 max pixels per img & 2000$\times$28$\times$28 \\
        SFT Stage 2 alignment weight & 2.0 \\
        SFT Stage 3 alignment weight & 2.0 \\
        \bottomrule
    \end{tabular}%
    }
\end{table}

\textbf{Latent only backpropagation in SFT Stage 2.}
To ensure that gradients flow only through the latent representations to the model parameters, we replace $\mathcal{L}_{\text{align-obs}}$ with the following surrogate loss:
\begin{equation}
   \mathcal{L'}_{\text{align-obs}} = \frac{1}{N}\sum_i \text{stop\_grad}(\frac{\partial \mathcal{L}_{\text{align-obs}}}{\partial {\hat{\mathbf{h}}^{(i, L)}_{\text{latent}}}})^\top {\hat{\mathbf{h}}^{(i, L)}_{\text{latent}}},
\end{equation}
where $\text{stop\_grad}$ denotes the stop-gradient operation, $\hat{\mathbf{h}}^{(i, L)}_{\text{latent}}$ are the generated latent embeddings in the student CoT, and $L$ is the number of layers in the MLLM’s language model. By differentiating $\mathcal{L'}_{\text{align-obs}}$ with respect to the model parameters $\theta$:
\begin{equation}
    \frac{\partial \mathcal{L'}_{\text{align-obs}}}{\partial \theta}= \frac{\partial \mathcal{L}_{\text{align-obs}}}{\partial {\hat{\mathbf{h}}^{(i, L)}_{\text{latent}}}} \frac{\partial {\hat{\mathbf{h}}^{(i, L)}_{\text{latent}}} }{\partial \theta},
\end{equation}
we can see that minimizing $\mathcal{L'}_{\text{align-obs}}$ is equivalent to optimizing only the generated latent embeddings.

\subsection{RL Training}
We implement our RL training on EasyR1 \citep{zheng2025easyr1}, an open-source RL training framework for multimodal LLMs.

\textbf{Hyperparameters.}  We summarize the RL training hyperparameters in Table \ref{tab:hyperparameters_rl}. Note that we apply an “accuracy threshold’’ to retain only samples whose group accuracy falls below this threshold but remains non-zero, preventing the advantage from vanishing when the responses in a group are all correct or incorrect.
\begin{table}[htbp]
    \centering
    \caption{\textbf{Hyperparameters for RL.} }
    \label{tab:hyperparameters_rl}
    \resizebox{0.33\textwidth}{!}{%
    \begin{tabular}{lc}
        \toprule 
        Hyperparameter & Value \\
        \midrule 
        learning rate & 0.000001 \\
        batch size & 64 \\
        weight decay & 0.01 \\
        rollout size & 8 \\
        temperature & 0.5 \\
        max response length & 4096 \\
        VLPO $\sigma$ & 10.0 \\
        SFT Stage 1 steps & 3885 (3 epochs) \\
        SFT Stage 2 steps & 1000 (1.03 epoch) \\
        SFT Stage 3 steps & 1000 (1.03 epoch) \\
        latent size & 10 \\
        max pixels per img & 2000$\times$28$\times$28 \\
        accuracy threshold & 0.6 \\
        \bottomrule
    \end{tabular}%
    }
\end{table}

\subsection{Training Data Construction}
\label{app:data_construction}
We provide the prompt for the API models used in the \textit{stage 3} of our training data construction pipeline as below:

\begin{promptbox}{System Prompt for Training Data Construction}
You are a helpful assistant. Your task is to identify which parts of the solution rely on visual observations (marked by \verb|<latent>|\verb|</latent>|). Wrap these essential observations in \verb|<observation>|...\verb|/<observation>| while leaving all other text unchanged. The input contains only a question (with conditions) and a solution chain-of-thought text — no images. Identifying observations does not require access to images. Segments are delimited by \verb|<STEP_i>| ... \verb|<END_STEP_i>|.\\
\\
Rules:\\
1. Do NOT remove, rename, or reorder any markers. Always output all markers exactly as given.\\
2. Do NOT add new content to the reasoning.\\
3. Inside \verb|<observation>|...\verb|</observation>|, include as little as possible — only the key tokens that truly require image observation.\\
4. Do NOT put information that can be inferred from the question text or earlier reasoning into \verb|<observation>|...\verb|</observation>|.\\
5. If you think there's no observation, just return the input text unchanged.\\
\end{promptbox}

\section{Detailed Experimental Setup }
\label{app:detailed_experimental_setup}
We use the VLMEvalKit \citep{duan2024vlmevalkit} framework for our evaluation.
We set the inference-time maximum visual tokens to be 8192$\times$28$\times$28. 
Below is the system prompt we used for evaluation. 

\begin{promptbox}{System Prompt for Evaluation}
You are an expert multimodal large language model designed to reason with latent visual embeddings.
\end{promptbox}

To ensure reliable evaluation, we first apply a rule-based judge (exact match) and then employ DeepSeek-V3.1 or Gemini-2.5-Pro as secondary judges.

\section{Case Studies}
\subsection{Monet-SFT-125K Examples}
\begin{figure*}[t]
\centering
\begin{tcolorbox}[title=\textbf{Monet-SFT-125K Example 1: Cropping the Crucial Region (from Zebra-CoT Visual Search)},
    colback=gray!5, colframe=gray!40,
    left=1mm, right=1mm, top=1mm, bottom=1mm,
    enhanced, sharp corners, width=\textwidth]

\textbf{Question:} According to the survey presented in the infographic, what percentage of parents in the U.S. delay the Varicella vaccine for their children beyond the CDC's recommendations?\\Put your final answer within \verb|\\boxed{}|."
 \\[4pt]

\textbf{Question Image:}\\
\includegraphics[width=0.99\textwidth]{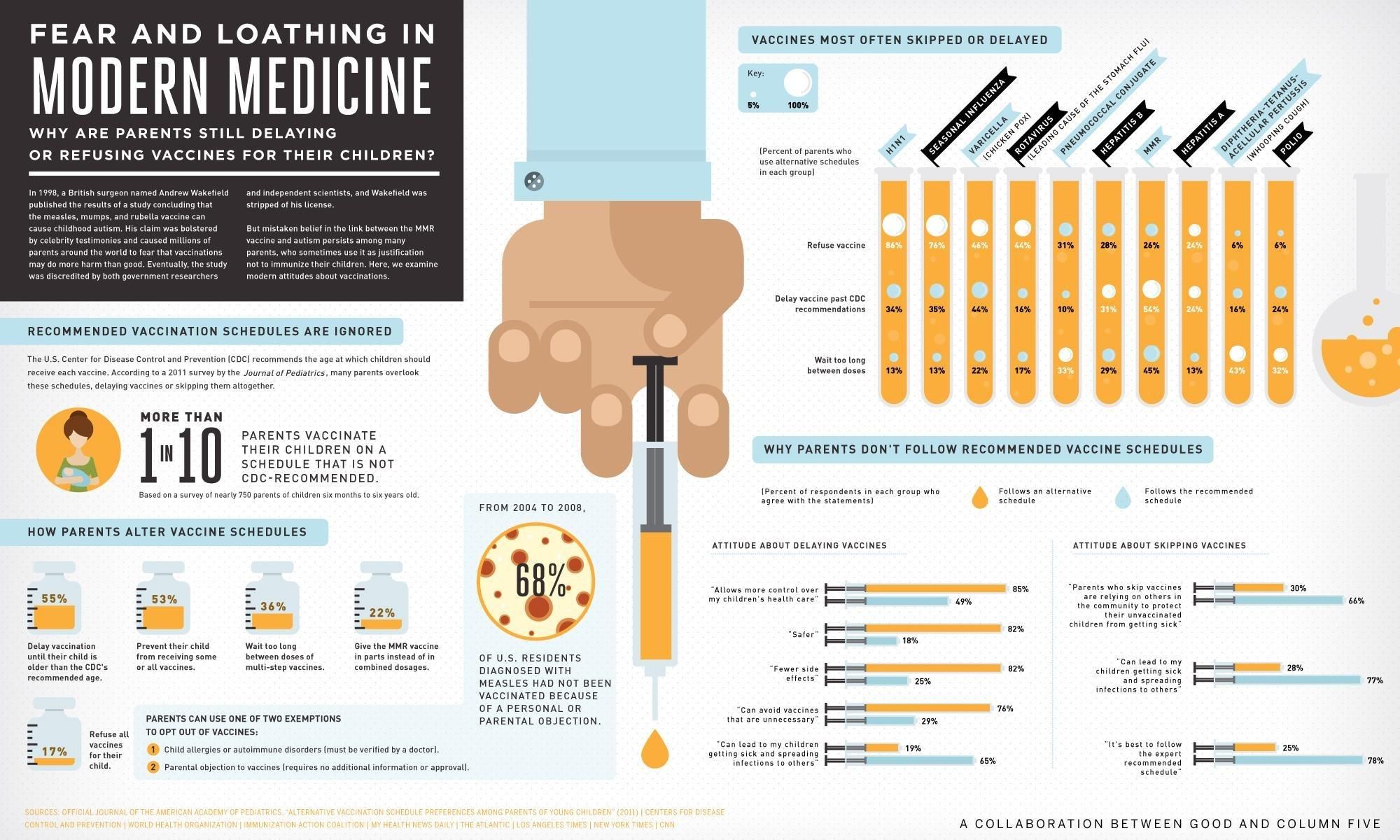} \\[4pt]

\textbf{Image-text Interleaved CoT:}\\
The question asks for the percentage of parents who delay the Varicella vaccine according to CDC recommendations, based on the provided infographic. I need to find the section of the infographic that lists vaccines and the percentages for delaying them past CDC recommendations. I will focus on the Varicella vaccine data.

\includegraphics[width=0.12\textwidth]{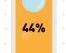} \\[4pt]

The cropped image shows the specific data point for the \verb|<observation>|Varicella vaccine\verb|</observation>| under the \"Delay vaccine past CDC recommendations\" row. Therefore, the final answer is \boxed{44\%}.

\end{tcolorbox}
\caption{\textbf{Example of Monet-SFT-125K: cropping the crucial region.}}
\label{fig:training_data_example1}
\end{figure*}

\begin{figure*}[t]
\centering
\begin{tcolorbox}[title=\textbf{Monet-SFT-125K Example 2: Creating New Visual States (from Zebra-CoT 3D Object Counting)},
    colback=gray!5, colframe=gray!40,
    left=1mm, right=1mm, top=1mm, bottom=1mm,
    enhanced, sharp corners, width=\textwidth]

\textbf{Question:} Consider the initial arrangement of objects presented in the image below. You are required to perform a series of removal operations in the specified order:\\Step 1: Eliminate all objects identified as PYRAMIDs.\\Step 2: Eliminate all objects that are BLUE in color.\\Step 3: Eliminate all objects that are RED in color.\\After completing these steps, determine the total count of orange objects remaining.\\Put your final answer within \verb|\\boxed{}|."
 \\[4pt]

\textbf{Question Image:}\\
\includegraphics[width=0.27\textwidth]{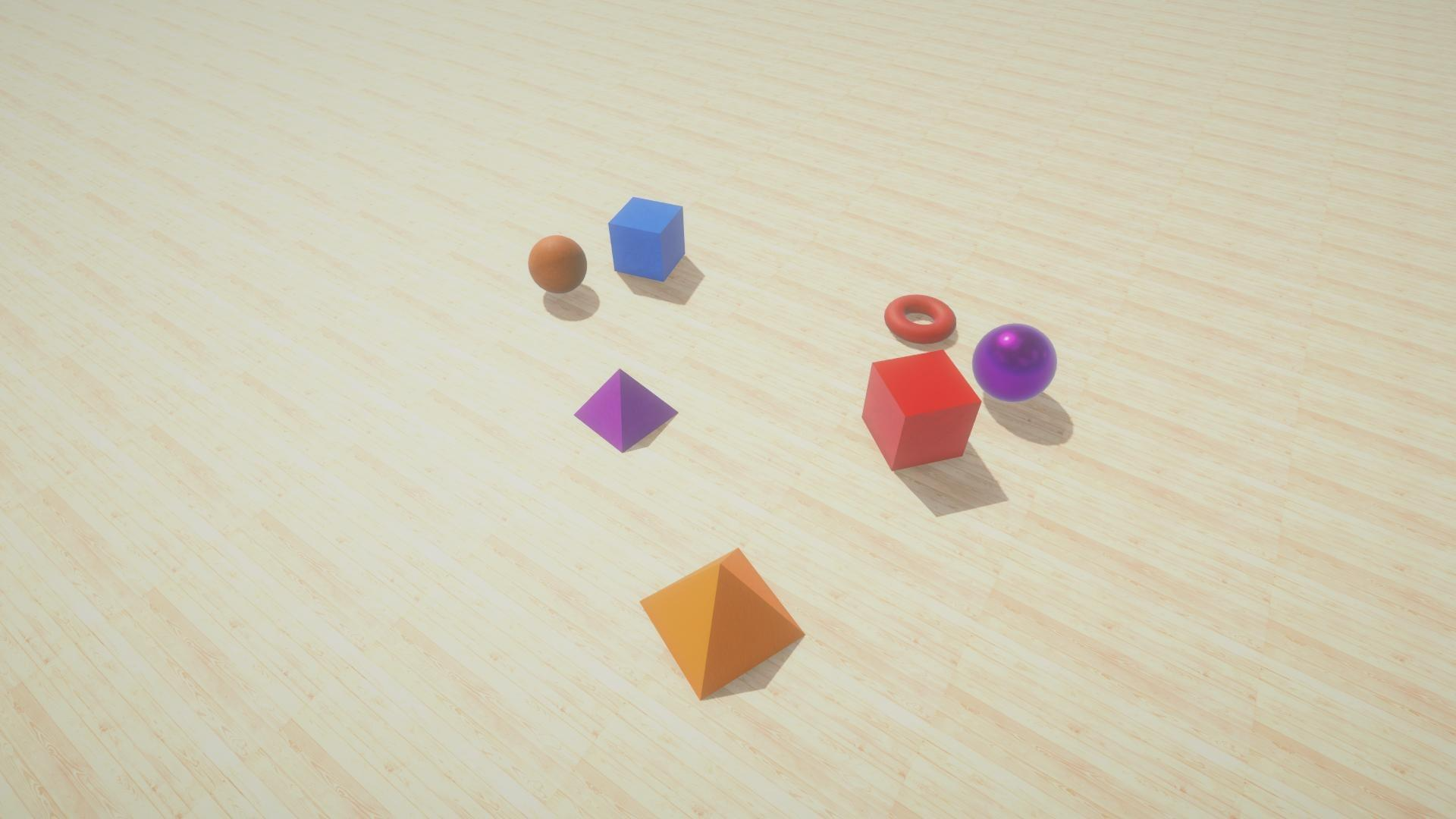} \\[4pt]

\textbf{Image-text Interleaved CoT:}\\
I will begin by analyzing the initial set of objects shown in the problem image. There are \verb|<observation>|7\verb|</observation>| distinct objects in the initial configuration. My approach will be to systematically apply each removal step and visually track the remaining objects before arriving at the final count.
The first operation requires the removal of all pyramid-shaped objects from the scene. This means both the \verb|<observation>|purple pyramid\verb|</observation>| and the \verb|<observation>|orange pyramid\verb|</observation>| will be taken away. To visualize the updated set of objects, I will generate an image reflecting this change.\\[4pt]
\includegraphics[width=0.27\textwidth]{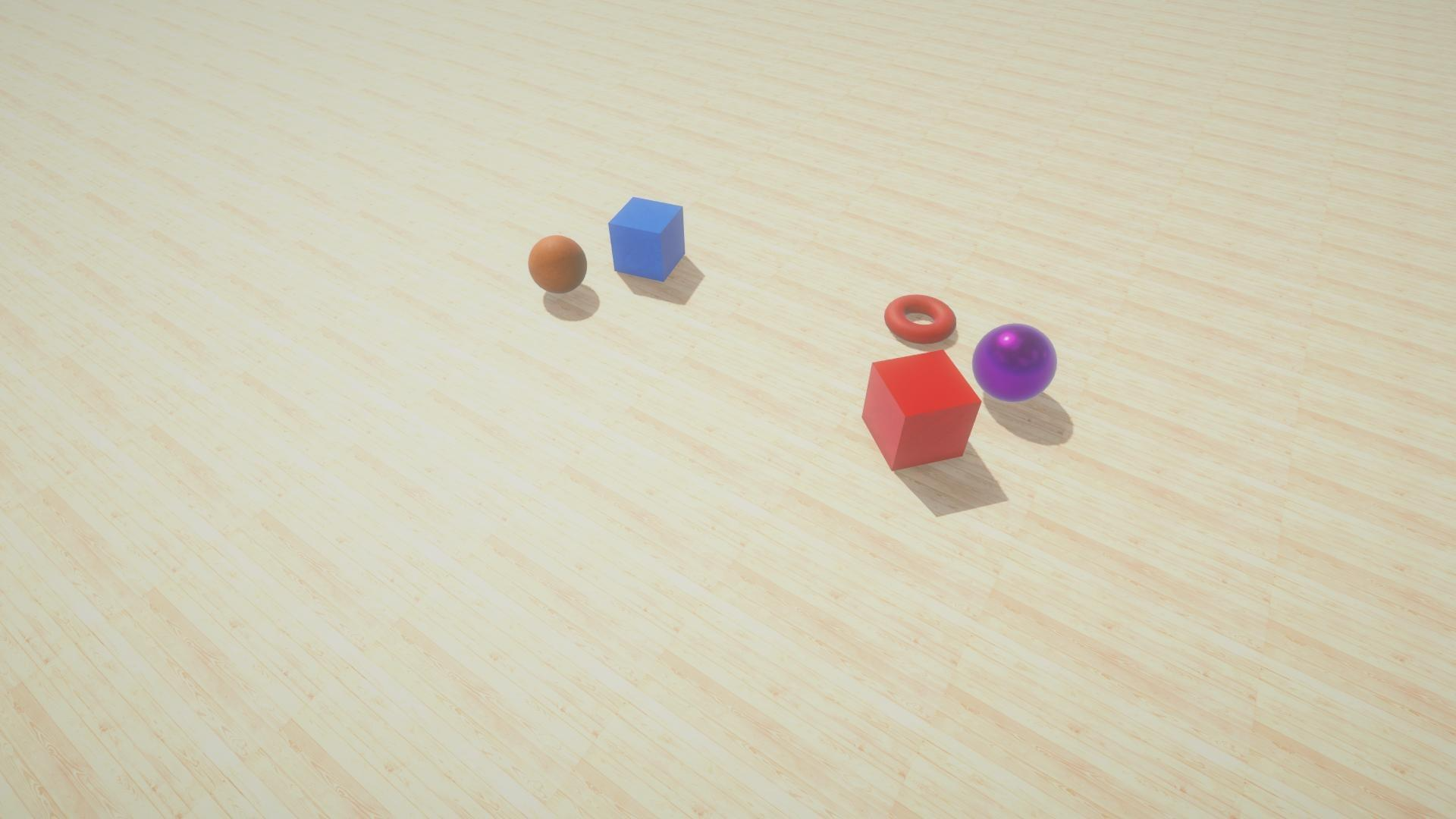} \\[4pt]
After removing the pyramids, as seen in the previous visual, we are left with \verb|<observation>|five\verb|</observation>| objects: an \verb|<observation>|orange sphere\verb|</observation>|, a \verb|<observation>|blue cube\verb|</observation>|, a \verb|<observation>|red torus\verb|</observation>|, a \verb|<observation>|red cube\verb|</observation>|, and a \verb|<observation>|purple sphere\verb|</observation>|. The next instruction is to remove all objects that are blue. This specifically targets the \verb|<observation>|blue cube\verb|</observation>|. I will generate a new visual to show the state after this removal.\\[4pt]
\includegraphics[width=0.27\textwidth]{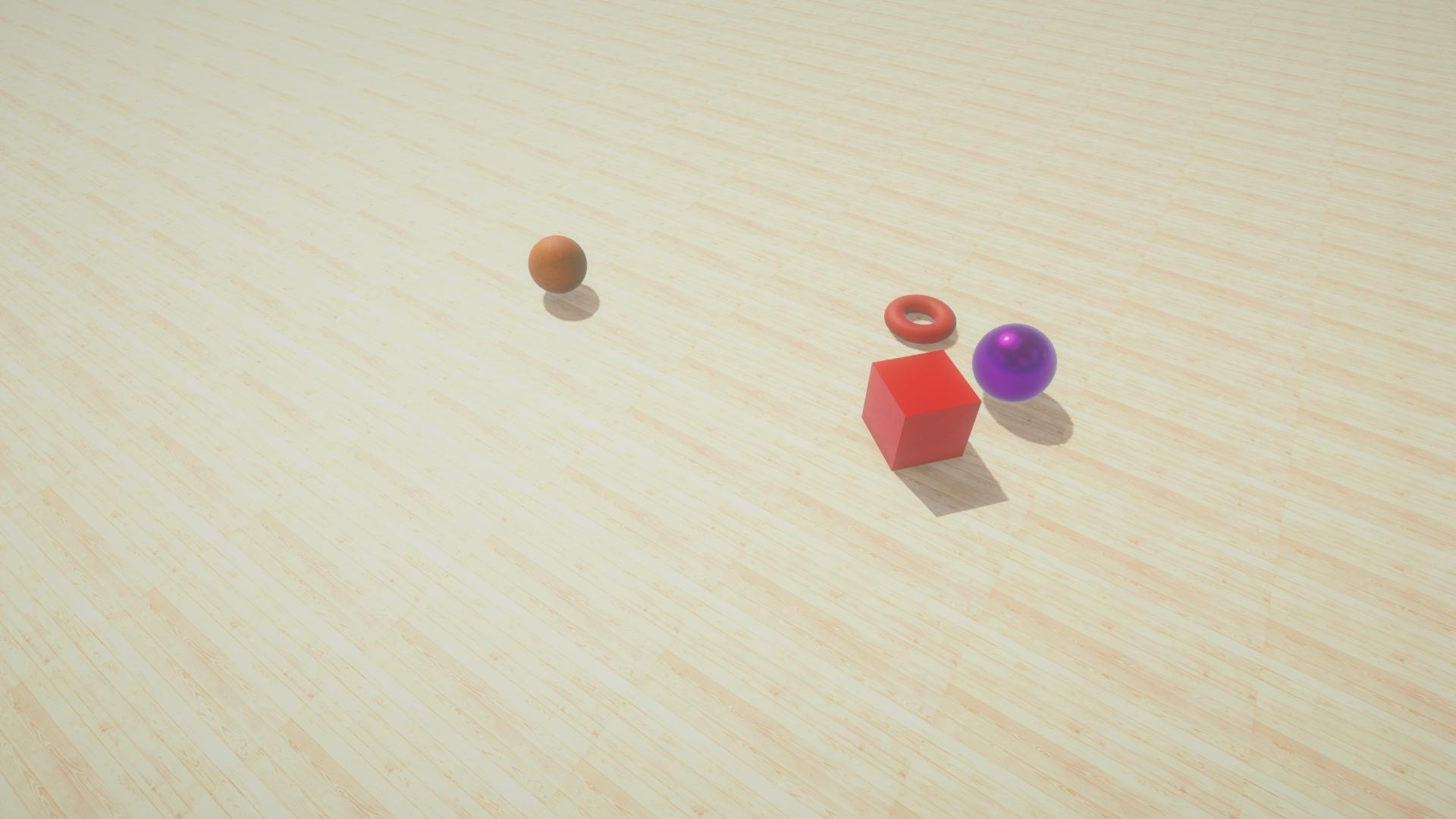} \\[4pt]
...\\[4pt]
Having completed all specified removal operations, the final visual shows \verb|<observation>|two\verb|</observation>| objects remaining: an \verb|<observation>|orange sphere\verb|</observation>| and a \verb|<observation>|purple sphere\verb|</observation>|. The question asks for the count of orange objects. From the two remaining objects, only one is orange. The final answer is \boxed{1}. 
\end{tcolorbox}
\caption{\textbf{Example of Monet-SFT-125K: creating new visual states.}}
\label{fig:training_data_example2}
\end{figure*}

\begin{figure*}[t]
\centering
\begin{tcolorbox}[title=\textbf{Monet-SFT-125K Example 3: Drawing Auxiliary Lines and Bounding Boxes (from CogCoM)},
    colback=gray!5, colframe=gray!40,
    left=1mm, right=1mm, top=1mm, bottom=1mm,
    enhanced, sharp corners, width=\textwidth]

\textbf{Question:}What is the sum of real users below 1 million?\\Put your final answer within \verb|\\boxed{}|."
 \\[4pt]

\textbf{Question Image:}\\
\includegraphics[width=0.3\textwidth]{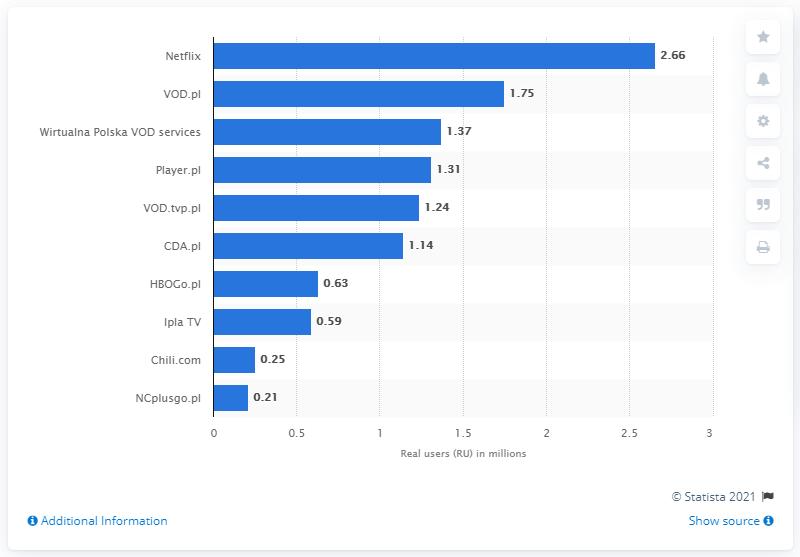} \\[4pt]

\textbf{Image-text Interleaved CoT:}\\
Draw a line to obtain the new image after the line is drawn.\\
\includegraphics[width=0.3\textwidth]{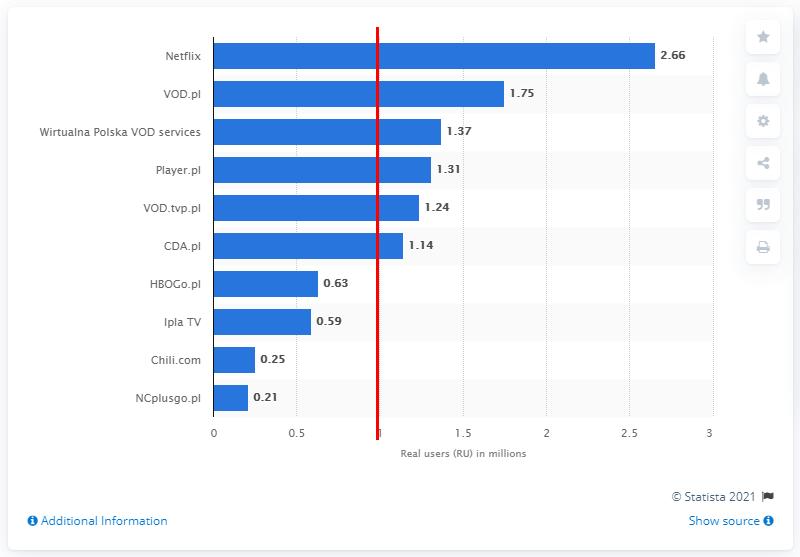} \\[4pt]
Outline each bar column, which are located at Netflix, VoD.pI, Wirtualna Polska VOD services, Player.pl, VoD.tvp.pI, CDA.pI, HBOGo.pI, lpla TV, Chili.com, NCplusgo.pl. \\[4pt]
\includegraphics[width=0.3\textwidth]{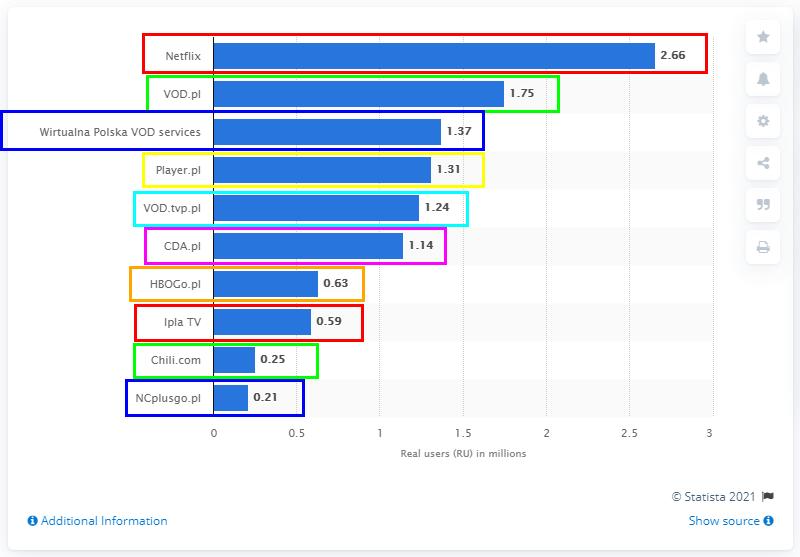} \\[4pt]
Analyze the value of each bar in relation to x=1. The values obtained from HBOGo.pI, lpla TV, Chili.com, and NCplusgo.pl are all \verb|<observation>|less than one million\verb|</observation>|, with respective values of \verb|<observation>|0.63, 0.59, 0.25, and 0.21.\verb|</observation>| Calculate their total sum as 0.63 + 0.59 + 0.25 + 0.21 = 1.68, so the answer is 1.68.
 The final answer is \boxed{1.68}. 
\end{tcolorbox}
\caption{\textbf{Example of Monet-SFT-125K: drawing auxiliary lines and bounding boxes.}}
\label{fig:training_data_example3}
\end{figure*}

We select three examples from Monet-SFT-125K to illustrate the data structure and corresponding visual operations. These examples cover three representative types of visual operations: cropping key regions, generating new visual states, and drawing auxiliary lines or bounding boxes. As shown in Figures \ref{fig:training_data_example1}, \ref{fig:training_data_example2}, and \ref{fig:training_data_example3}, each training instance contains an image–text interleaved CoT with intermediate auxiliary images that facilitate the visual reasoning process. Additionally, tokens that encode key observations from the auxiliary images are wrapped in \verb|<observation>| and \verb|</observation>|, which are used in the alignment loss during SFT Stage 2.

\subsection{Inference Examples}
We present some inference examples generated by Monet-7B in this section to show the generality of tasks that Monet-7B can solve. For clarity, we don't map the generated latent embeddings to the language space as this will produce meaningless tokens. Instead, we use \verb|<latent><latent_embeddings></latent>| to represent the latent reasoning segments.

\begin{figure*}[t]
\centering
\begin{tcolorbox}[title=\textbf{Monet-7B Inference Example 1: 3D Spatial Reasoning (from VisualPuzzles)},
    colback=gray!5, colframe=blue!60,
    left=1mm, right=1mm, top=1mm, bottom=1mm,
    enhanced, sharp corners, width=\textwidth]

\textbf{Question:}Given a pair of chairs arranged at a specific angle, select the option from the four provided that matches the angle of the chairs in the question.
Options: Choose from (A) (B) (C) (D) in the image.
"
 \\[4pt]

\textbf{Question Image:}\\
\includegraphics[width=0.6\textwidth]{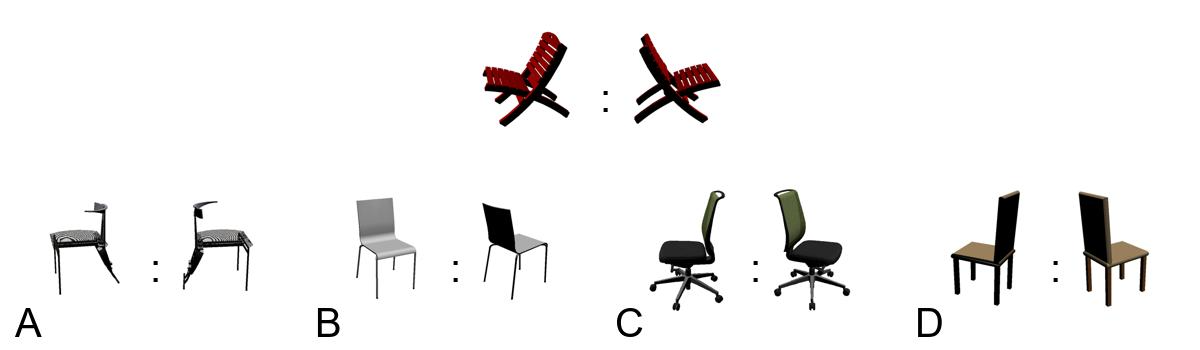} \\[4pt]

\textbf{Response by Monet-7B:}\\
The problem asks to identify the option that matches the given angle between the chairs. I will analyze the provided image to determine the angle between the chairs and then compare it with the given choices.To accurately assess the angle between the chairs, I will focus on the visual representation of the chairs in the problem image.
\verb|<latent><latent_embeddings></latent>|The zoomed-in view of the chairs in the problem image clearly shows the angle between them. This visual confirms the specific spatial relationship between the two chairs. Therefore, the final answer is \boxed{A}.

\end{tcolorbox}
\caption{\textbf{Inference example: 3D spatial reasoning.} Instead of describing the angles between the chairs in language, Monet directly reasons with latent embeddings before giving the final answer.}
\label{fig:inference_example1}
\end{figure*}

\begin{figure*}[t]
\centering
\begin{tcolorbox}[title=\textbf{Monet-7B Inference Example 2: 2D Transformation (from VisualPuzzles)},
    colback=gray!5, colframe=blue!60,
    left=1mm, right=1mm, top=1mm, bottom=1mm,
    enhanced, sharp corners, width=\textwidth]

\textbf{Question:} The number 4 is reflected twice in \verb|<image1>|. Which among the options in \verb|<image2>| should appear in the field with the question mark if we do the same with the number 5?
"
 \\[4pt]

\textbf{Question Image:}\\
\includegraphics[width=0.5\textwidth]{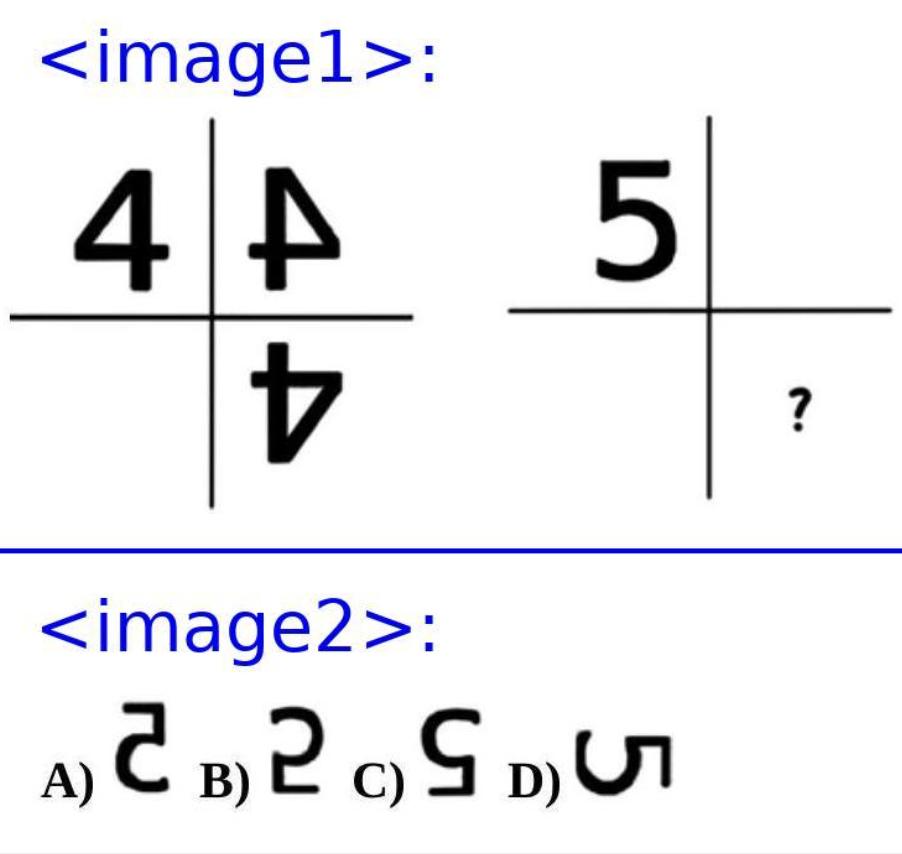} \\[4pt]

\textbf{Response by Monet-7B:}\\
To determine the correct answer, we need to identify the pattern of reflection for the number 5, similar to how the number 4 is reflected in the provided image.To better understand the reflection pattern, I will generate a zoomed-in view of the relevant section of the image.
\verb|<latent><latent_embeddings></latent>|The zoomed-in image clearly shows how number 4 is reflected. The reflection pattern for the number 4 is obtained by flipping it twice. Once horizontally, once vertically. Therefore, the final answer is \boxed{C}.

\end{tcolorbox}
\caption{\textbf{Inference example: 2D transformation.} By generating latent embeddings, Monet successfully identifies the flipping rule of the number.}
\label{fig:inference_example2}
\end{figure*}

\begin{figure*}[t]
\centering
\begin{tcolorbox}[title=\textbf{Monet-7B Inference Example 3: Complex Diagram Reasoning (from MME-RealWorld-Lite)},
    colback=gray!5, colframe=blue!60,
    left=1mm, right=1mm, top=1mm, bottom=1mm,
    enhanced, sharp corners, width=\textwidth]

\textbf{Question:} Which country is the top sales country? The choices are listed below:\\
(A) Canada\\
(B) Germany\\
(C) UK\\
(D) United States\\
(E) The image does not feature the number.
 \\[4pt]

\textbf{Question Image:}\\
\includegraphics[width=0.99\textwidth]{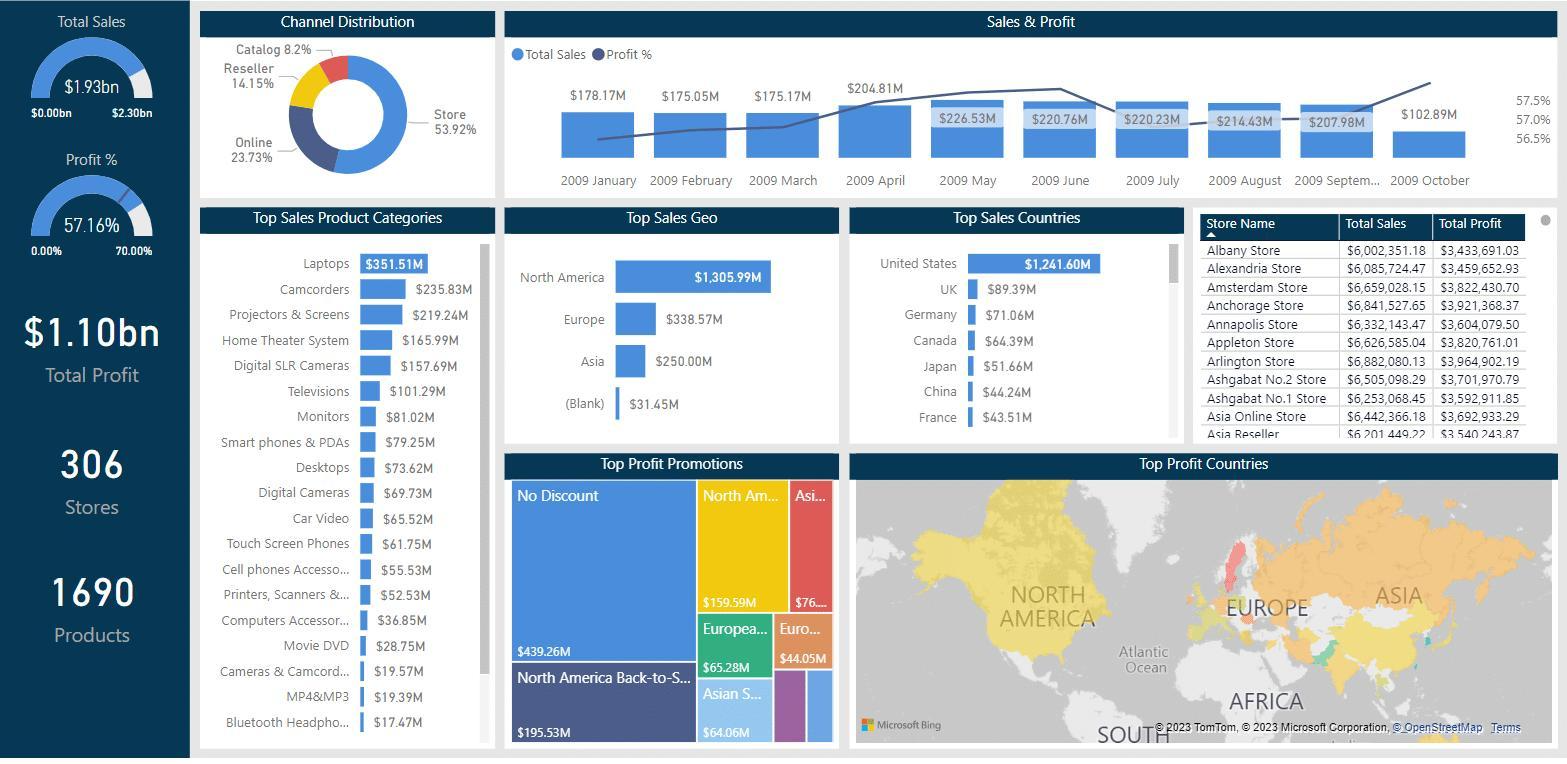} \\[4pt]

\textbf{Response by Monet-7B:}\\
To determine the top sales country, I need to locate the section of the image that displays sales by country.
\verb|<latent><latent_embeddings></latent>|The highlighted area in the image clearly shows the "Top Sales Countries" section. It lists various countries along with their respective sales figures. The sales figures are in millions of dollars.
From the highlighted section, I can see that the United States has the highest sales at 1,241.60M, followed by the UK, Germany, and other countries. This directly answers the question about the top sales country. The final answer is \boxed{D}.

\end{tcolorbox}
\caption{\textbf{Inference example: complex diagram reasoning.} Monet-7B exhibit a \textbf{hierarchical} reasoning pattern. Firstlt, it focus on the relevant section in the image by generating latent embeddings: ``\textit{The highlighted area in the image clearly shows the ``Top Sales Countries" section.}'' Then, it accurately identifies the contents in the ``Top Sales Countries'' section and gives the correct answer.}
\label{fig:inference_example3}
\end{figure*}

\begin{figure*}[t]
\centering
\begin{tcolorbox}[title=\textbf{Monet-7B Inference Example 4: Commonsense QA (from VisualPuzzles)},
    colback=gray!5, colframe=blue!60,
    left=1mm, right=1mm, top=1mm, bottom=1mm,
    enhanced, sharp corners, width=\textwidth]

\textbf{Question:} Based on the relationship between Origin and Analogy, select the option from the candidates that best completes this analogy.\\
Options: Choose from (A) (B) (C) (D) in the image.
 \\[4pt]

\textbf{Question Image:}\\
\includegraphics[width=0.99\textwidth]{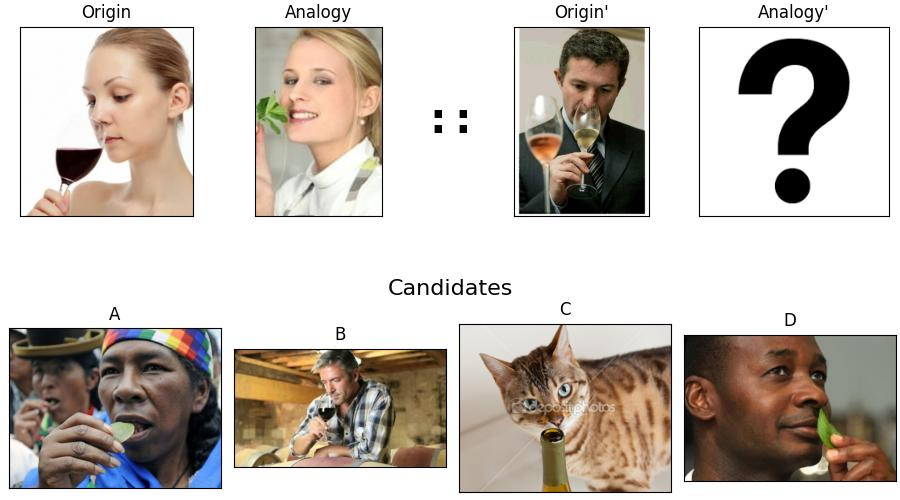} \\[4pt]

\textbf{Response by Monet-7B:}\\
To identify the correct image that completes the analogy, I will analyze the relationship between the "Origin" and "Analogy" images in the provided problem image. The goal is to find a similar relationship between the "Origin'" image and the missing "Analogy'" image. To better understand the relationship, I will focus on the "Origin" and "Analogy" images to identify the specific action being performed by the person in each image. \verb|<latent><latent_embeddings></latent>| \\
The zoomed-in view of the "Origin" image clearly shows  a person  drinking wine, while the "Analogy" image depicts  a person  smelling a flower. This indicates that the action involves the sense of smell.Therefore, the final answer is \boxed{D}.

\end{tcolorbox}
\caption{\textbf{Inference example: commonsense QA.} Monet-7B correctly identifies the connection between the ``Origin'' and the ``Analogy'' image, showing its commonsense reasoning capability.}
\label{fig:inference_example4}
\end{figure*}

\begin{figure*}[t]
\centering
\begin{tcolorbox}[title=\textbf{Monet-7B Inference Example 5: Fine-grained OCR (from MME-RealWorld-Lite)},
    colback=gray!5, colframe=blue!60,
    left=1mm, right=1mm, top=1mm, bottom=1mm,
    enhanced, sharp corners, width=\textwidth]

\textbf{Question:} Where does Joseph work? The choices are listed below:\\
(A) Boston.\\
(B) Miami.\\
(C) Chicago.\\
(D) New York.\\
(E) The image does not feature the related information.
 \\[4pt]

\textbf{Question Image:}\\
\includegraphics[width=0.7\textwidth]{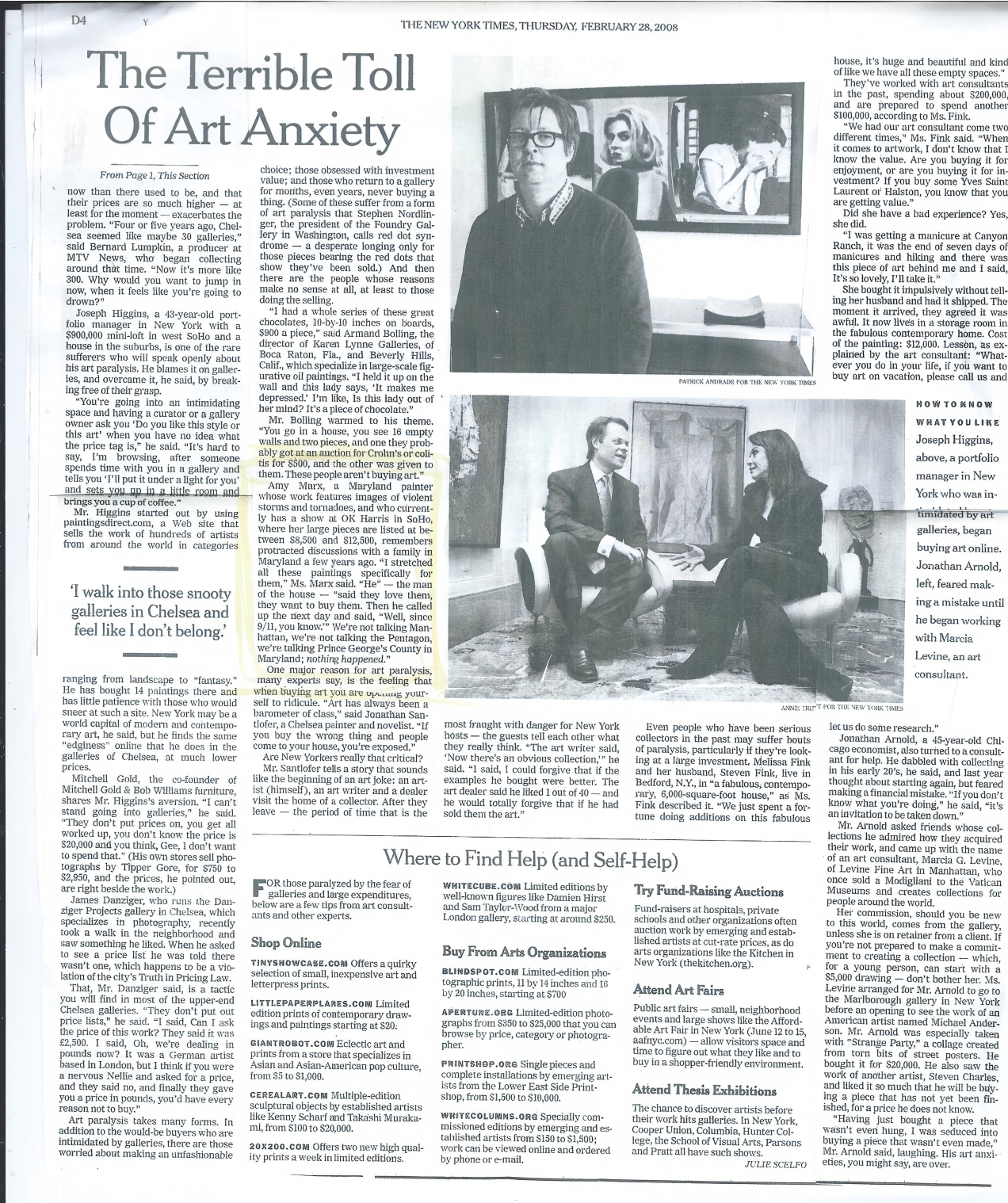} \\[4pt]

\textbf{Response by Monet-7B:}\\
To determine where Joseph works, I need to locate his name or relevant information within the article. I will zoom in on the section of the article that mentions Joseph to clearly identify his location. \verb|<latent><latent_embeddings></latent>| \\
The zoomed-in image clearly shows that Joseph Higgins is a portfolio manager in  New York. This directly answers the question about where Joseph works.Therefore, the final answer is \boxed{D}.

\end{tcolorbox}
\caption{\textbf{Inference example: fine-grained OCR.} Monet-7B accurately identifies the key information, which is located in the middle of the rightmost region of the image.}
\label{fig:inference_example5}
\end{figure*}

The tasks of the selected examples include 3D Spatial Reasoning (Figure \ref{fig:inference_example1}), 2D transformation (Figure \ref{fig:inference_example2}), complex diagram reasoning (Figure \ref{fig:inference_example3}), commonsense QA (Figrue \ref{fig:inference_example4}), fine-grained OCR (Figure \ref{fig:inference_example5}), and math reasoning (Figure \ref{fig:inference_example6}). Note that Monet not always uses latent thinking. For example, for the pure-text math problem in Figure \ref{fig:inference_example6}, and Monet directly rely text-based reasoning to solve the problem.

\begin{figure*}[t]
\centering
\begin{tcolorbox}[title=\textbf{Monet-7B Inference Example 6: Math Reasoning (from VisualPuzzles)},
    colback=gray!5, colframe=blue!60,
    left=1mm, right=1mm, top=1mm, bottom=1mm,
    enhanced, sharp corners, width=\textwidth]

\textbf{Question:} Every night the wizard Tilim makes the weather forecast for the king. When Tilim gets it right he gets 3 gold coins, but when he makes a mistake, he pays a fine of 2 gold coins. After making the prediction for 5 days, Tilim did the math and discovered that he neither won nor lost coins. How many times did he get the weather forecast right in those 5 days?\\                             Options:\\                                                                                                                                (A) 1\\
(B) 2\\
(C) 3\\
(D) 4\\[4pt]

\textbf{Response by Monet-7B:}\\
To determine the number of correct weather forecasts, we can set up an equation based on the given conditions. Let's define:               - \( x \) as the number of correct forecasts (where he earns 3 gold coins per correct forecast).\\
- \( y \) as the number of incorrect forecasts (where he loses 2 gold coins per incorrect forecast)\\
The wizard made a total of 5 forecasts, so:\\
\[ x + y = 5 \]\\
He neither won nor lost any gold, which means the total gain from correct forecasts equals the total loss from incorrect forecasts: \[ 3x = 2y \]We now have a system of two equations:\\
1. \( x + y = 5 \)\\
2. \( 3x = 2y \)\\
We can solve this system by expressing \( y \) in terms of \( x \) from the first equation:\[ y = 5 - x \]Next, substitute \( y = 5 - x \) into the second equation:\[ 3x = 2(5 - x) \]Solve for \( x \):\[ 3x = 10 - 2x \]\[ 3x + 2x = 10 \]\[ 5x = 10 \]\[ x = 2 \]So, Tilim made 2 correct forecasts. To verify, we calculate \( y \):\[ y = 5 - x = 5 - 2 = 3 \]Now, check the total gold:\\
- Correct forecasts: \( 2 \times 3 = 6 \) gold\\
- Incorrect forecasts: \( 3 \times (-2) = -6 \) gold\\
- Total gold: \( 6 + (-6) = 0 \) gold. The calculations confirm that Tilim neither won nor lost any gold. Therefore, the number of correct forecasts is \boxed{2}. The final answer is \boxed{B}.

\end{tcolorbox}
\caption{\textbf{Inference example: math reasoning.} Monet-7B can perform pure-text math reasoning. Since the problem is not visual-related, it doesn't activate latent thinking mode.}
\label{fig:inference_example6}
\end{figure*}

\end{document}